\documentclass[10pt,twocolumn,letterpaper]{article}

\usepackage{iccv}
\usepackage{times}
\usepackage{epsfig}
\usepackage{graphicx}
\usepackage{amsmath}
\usepackage{amssymb}

\usepackage[accsupp]{axessibility}  % Improves PDF readability for those with disabilities.

\usepackage[dvipsnames]{xcolor}
\usepackage[toc,page]{appendix}

\definecolor{myorange}{RGB}{255,69,0}

\usepackage[pagebackref=true,breaklinks=true,letterpaper=true,colorlinks,bookmarks=false]{hyperref}

\usepackage{xspace}
\usepackage{booktabs}

\usepackage{subcaption}

\makeatletter
\DeclareRobustCommand\onedot{\futurelet\@let@token\@onedot}
\def\@onedot{\ifx\@let@token.\else.\null\fi\xspace}

\def\eg{\emph{e.g}\onedot} 
\def\ie{\emph{i.e}\onedot}

\def\etal{\emph{et al}\onedot}
\makeatother

\newcommand{\sect}[1]{Section~\ref{#1}}

\newcommand{\eqn}[1]{Equation~(\ref{#1})}
\newcommand{\fig}[1]{Figure~\ref{#1}}
\newcommand{\tbl}[1]{Table~\ref{#1}}

\newcommand{\ignore}[1]{}

\newcommand{\lsdata}{V-HICO\xspace}

\definecolor{rowblue}{RGB}{220,230,240}
\definecolor{myorchid}{RGB}{150,10,30}
\definecolor{myblue}{RGB}{10,30,250}
\definecolor{mygreen}{RGB}{10,120,10}

\def\params{\theta}

\def\losssparsity{\mathcal{L}_{\text{spa}}}
\def\lossbinary{\mathcal{L}_{\text{cls}}}

\def\wordembedding{e}
\def\dictembedding{\mathcal{E}}

\def\attention{\Phi}

\def\losscontrastive{\mathcal{L}_{\text{c}}}
\def\featuresnegative{\mathcal{F}}

\def\losscontrastive{\mathcal{L}_{C}}
\def\lossspatiotemporal{\mathcal{L}_{\textrm{ST}}}
\def\losslanguage{\mathcal{L}_{L}}
\def\losstemporal{\mathcal{L}_{T}}

\iccvfinalcopy % *** Uncomment this line for the final submission

\def\iccvPaperID{5475} % *** Enter the ICCV Paper ID here

\ificcvfinal\fi

\begin{document}

%%%%%%%%% TITLE
\title{Weakly Supervised Human-Object Interaction Detection in Video via Contrastive Spatiotemporal Regions}
%\title{Weakly Supervised Human and Object Detection \\ via Spatiotemporal Interactions}

% {\tt\small lishuang@mit.edu}
% {\tt\small yilundu@mit.edu}
% {\tt\small torralba@mit.edu}
% {\tt\small Josef.Sivic@ens.fr}
% {\tt\small brussell@adobe.com}

% \author{Shuang Li \\
% MIT \\
% \and
% Yilun Du\\
% MIT \\
% \and
% Antonio Torralba \\
% MIT \\
% \and
% Josef Sivic \\
% CIIRC CTU \footnote[1]{test} \\
% \and
% Bryan Russell \\
% Adobe \\
% }

\author{
Shuang Li$^{1}$$^*$$\;\;\;$
Yilun Du$^{1}$ $\;\;\;$ 
Antonio Torralba$^{1}$ $\;\;\;$ 
Josef Sivic$^2$\;\;
Bryan Russell$^{3}$\;\; \\
$^1$MIT\;\;\;  $^2$CIIRC CTU\;\;\;  $^3$Adobe \\
\href{https://shuangli-project.github.io/weakly-supervised-human-object-detection-video/}{\small{https://shuangli-project.github.io/weakly-supervised-human-object-detection-video}}
}

\maketitle
\ificcvfinal\thispagestyle{empty}\fi

\let\thefootnote\relax\footnote{$^2$Czech Institute of Informatics, Robotics and Cybernetics at the Czech
Technical University in Prague}
\let\thefootnote\relax\footnote{$^*$Work done at Adobe Research during SL's summer internship}

\vspace{-15pt}
\begin{abstract}
We introduce the task of weakly supervised learning for detecting human and object interactions in videos. Our task poses unique challenges as a system does not know what types of human-object interactions are present in a video or the actual spatiotemporal location of the human and the object. To address these challenges, we introduce a contrastive weakly supervised training loss that aims to jointly associate spatiotemporal regions in a video with an action and object vocabulary and encourage temporal continuity of the visual appearance of moving objects as a form of self-supervision. To train our model, we introduce a dataset comprising over 6.5k videos with human-object interaction annotations that have been semi-automatically curated from sentence captions associated with the videos. We demonstrate improved performance over weakly supervised baselines adapted to our task on our video dataset. 
\end{abstract}

\vspace{-2pt}
\section{Introduction}
\begin{figure}[t]
\begin{center}
\includegraphics[width=0.98\linewidth]{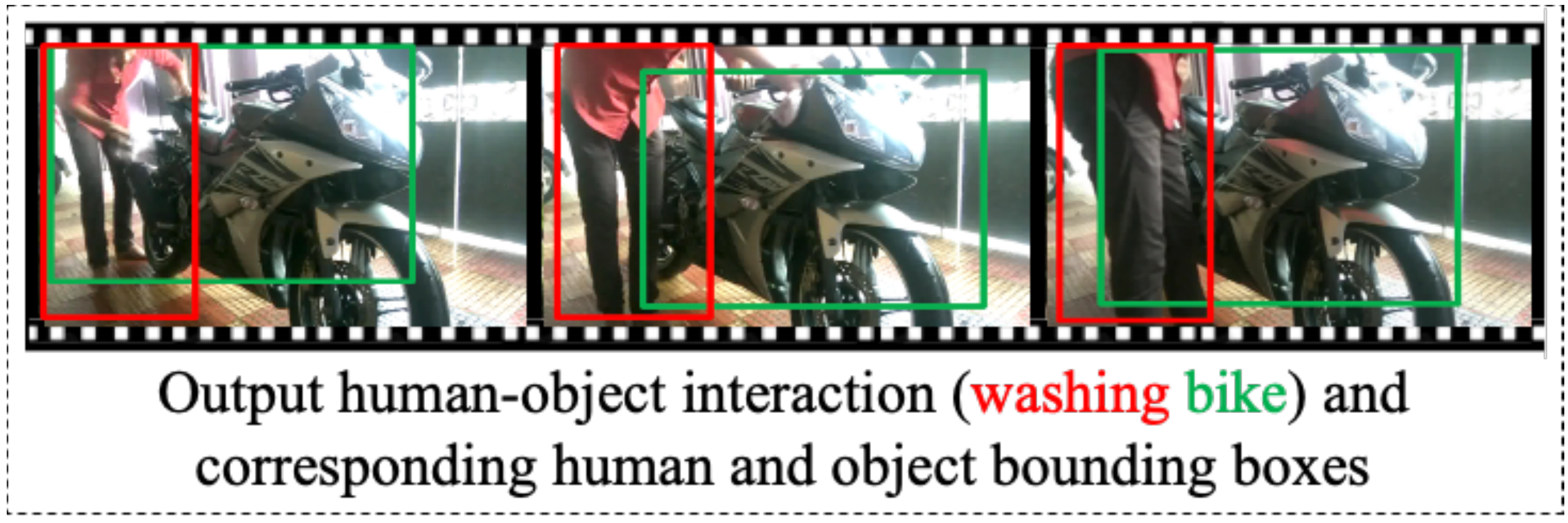}
\end{center}
\vspace{-15pt}
\caption{\small We seek to detect human-object interactions in videos. In this example, our system is able to detect ``human washing bike'' in the given video. 
Our approach learns to detect such interactions in a weakly supervised fashion, \ie, {without} requiring bounding box annotations at training time. (Video credit: Dude Chennai~\cite{urlfig1})
}
\label{fig:teaser_video}
\vspace{-15pt}
\end{figure}

\vspace{-2pt}
In this paper, we study the problem of weakly supervised human-object interaction detection in videos. 
%More specifically, a ideal system should detect relations (verb-object pairs) between humans and objects in videos given weak supervision in the form of associated verb-object keywords during training.
Given a video sequence, as illustrated in Figure~\ref{fig:teaser_video}, a system must correctly identify and localize the person and interacted object (``bike’’) in the scene, in addition to identifying the action (``washing’’) taken by the human, for the duration of the interaction in the video without bounding box supervision. 
% Achieving such a goal is challenging as we do not know what type of human-object interaction is present in the video and what are the spatiotemporal locations of humans and objects that depict the undergoing interactions.
While there has been an impressive progress in learning visual-language representations~\cite{radford2021learning,jia2021scaling,miech2020end} from hundreds of millions of captioned images or videos recently, the learnt representations focus on classifying or retrieving entire images or videos given a language query. Our task is more challenging as it requires the models to correctly detect both the human and object bounding boxes in multiple frames of the video.

% has many important applications, such as video retrieval, surveillance or human-robot interaction. 

Human-object interaction detection has been primarily studied in the context of still images~\cite{chao2018learning,chao2015hico,gupta2015visual,hou2020visual,wang2020discovering,wang2019deep,qi2018learning,xu2019learning,zhou2019relation,wan2019pose,song2020novel}.
% have focused mainly on still images. 
However, they are naturally temporal events that take place over a period of time. 
Interactions such as ``drinking’’ or ``pushing’’ occur between a human and an object over time, making videos a natural modality for studying this problem. 
% symbolize temporal interaction between human and object, making videos a more natural way to study this problem.

% Our task is challenging as we do not know the correspondence between the verb-object queries and spatiotemporal regions in the training videos. A system must learn to establish these correspondences without spatial bounding box supervision. 
Existing video-based methods primarily rely on strong bounding box supervision and having access to a fully annotated video dataset. 
However, relying on strong supervision has significant drawbacks.  
% Most existing human-object interaction approaches \cite{chao2018learning,gao2018ican,gupta2015visual} are further typically fully supervised, which suffers from several significant drawbacks. 
First, exhaustively annotating the spatial location of objects in a video is time consuming given the large number of frames in a video. 
Second, scaling to the large number of possible interactions and obtaining a sufficient number of ground truth bounding boxes is challenging due to the potentially open vocabulary of objects and actions and the combinatorial nature of human-object interactions.
Third, interactions typically follow a long-tailed distribution, with common human-object interactions occurring much more frequently than others \cite{yuhang2020novel,hou2020visual}. 
While supervised learning usually prefers common interactions, a robust human-object interaction detection system should instead perform equally well on both common and rare interactions.

In this work, we seek to leverage videos with verb and noun phrase annotations derived from natural language sentence captions to learn to detect human-object interactions in videos in a weakly supervised manner. Such an approach is advantageous as obtaining video-level annotations is significantly less costly than bounding boxes in videos. Leveraging such data makes it possible to scale training to a larger number of videos and vocabulary of objects and actions.

% As bounding box annotations are not available in captioned videos, our goal is to detect a human and an object interacting in a video in a weakly supervised fashion without the usage of bounding box annotations. 
% This goal is challenging as we do not know what type of human-object interaction is present in a given video and what are the spatiotemporal locations of humans and objects that depict the undergoing interactions.
% Moreover, in contrast to previous work detecting interactions with fixed vocabulary~\cite{Shang2017VideoVR,tsai2019GSTEG}, 
% We also detect unseen human-object interactions in the open vocabulary during testing.
% our method is able to leverage the pretrained language model to detect unseen human-object interactions in the open vocabulary during testing.
%
Our task is challenging as we do not know the correspondence between the verb-object queries and spatiotemporal regions in the training videos. A system must learn to establish these correspondences without spatial bounding box supervision. 
We thus propose a contrastive loss over spatiotemporal regions for detecting human-object interactions in videos. 
% To address the aforementioned correspondence challenge, we propose a contrastive loss over spatiotemporal regions for detecting human-object interactions in videos. 
%Our approach does not require spatial bounding box supervision. 
%As a result, our approach does not require bounding box annotations. 
% We thus propose a method with a contrastive weakly supervised spatiotemporal loss for this task. 
Our loss jointly associates candidate spatiotemporal regions with an action and object vocabulary in a weakly supervised manner and leverages cues about the temporal continuity of objects in motion as a form of self-supervision. 
% As a result, it does not require spatial bounding box supervision.
%Given weak supervision in the form of an action and object during training, our loss encourages the visual representation for likely human and object spatiotemporal region proposals to be close to the corresponding language embedding representation of the input action and object query. 
Such a formulation allows us to deal with an open vocabulary of language queries which is especially desirable in human-object interaction, due to the high prevalence of rare and unseen action and object combinations. 
% We show that such an approach enables us to detect interactions consisting of new unknown objects.

% We show that our formulation allows detecting rare and unseen action and object combinations.
%deal an open vocabulary of language queries with new unknown interactions. This is especially desirable in human-object interaction detection, due to the high prevalence of rare and unseen action and object combinations. 
% We show that such an approach enables us to detect interactions consisting of new unknown objects.
% The loss also maintains temporal continuity of the object's visual appearance throughout the video. 

% To enable human-object interaction detection in videos, we introduce a dataset of over 6k videos derived from the Moments in Time dataset~\cite{monfort2019moments} where human-object interactions have been semi-automatically curated from natural language captions accompanying the videos. 

Our paper has three main contributions: 
% (1) We present an approach that integrates spatiotemporal information for humans and objects for weakly supervised human-object interaction detection in videos. Thanks to a region attention module, our approach does not require manual bounding box annotations.
(1) We present an approach that integrates spatiotemporal information for humans and objects for weakly supervised human-object interaction detection in videos. Our approach does not require manual bounding box annotations. 
% (2) We present a contrastive loss over spatiotemporal regions that allows detecting rare and unseen human-object interactions. 
(2) We present a contrastive loss over spatiotemporal regions that leverages weak verb-object supervision from video captions and self-supervision from temporal continuity in video. 
It allows detecting rare and unseen human-object interactions in a zero-shot manner.
% It enables us to handle the human-object interactions with long-tailed distribution, and further allows us to detect rare and unseen human-object interactions in a zero-shot manner.
(3) We introduce a new dataset of over 6.5k videos to evaluate human-object interaction in videos. We demonstrate improved performance over weakly supervised baselines adapted to our task.
The dataset is made public to facilitate further research$^1$.
\let\thefootnote\relax\footnote{$^1$Code and dataset are available at 
\href{https://shuangli-project.github.io/VHICO-Dataset/}{https://shuangli-project.github.io/VHICO-Dataset}.}

\vspace{-3pt}
\section{Related work}
Closest to our approach is work in modelling video and natural language, visual relationship detection, and human-object interaction detection.

\textbf{Video and natural language.} 
Prior work has looked at jointly modeling video and natural language for tasks, such as captioning~\cite{krishna2017dense}, movie question answering~\cite{MovieQA}, and short clip retrieval~\cite{Rohrbach2015ADF,Xu2016MSRVTTAL}. 
More relevant are works that aim to more finely ``ground'' or align natural language in videos.  
Examples include retrieving moments from untrimmed videos~\cite{gao2017tall,hendricks2017localizing}, learning from video with aligned instructions~\cite{MiechZATLS2019,miech2020end}, and alignment of natural language with (spatio-) temporal regions in a video~\cite{Huang18}. 
Natural language poses hard challenges due to large open vocabulary and complex interactions due to composition.
% our effort is toward such reasoning.  

\textbf{Visual relationship detection.}
Previous work, \eg, \cite{baier2017improving,dai2017detecting,gkioxari2018detecting,gupta2015visual,han2018visual,liang2018visual,plesse2018visual,plummer2017phrase,sadeghi2011recognition,xu2019learning,yu2017visual,zhan2019exploring}, has studied detecting subject-predicate-object visual relations in single still images. 
% All of these approaches assume strong supervision where the triplet and accompanying subject/object bounding boxes are provided at training. 
This line of work has been extended to video with strong supervision~\cite{Shang2017VideoVR,tsai2019GSTEG}. 
% Also related are works to discover objects in video by reasoning through detected humans and actions~\cite{Yang19} or via natural language queries~\cite{zhou2018weakly}. However, these works do not explicitly model or evaluate the triplet relationship, and the former does not operate in an open-language setting. 
Closest to our approach is work on weakly supervised visual relationship detection \cite{peyre2017weakly,zhang2017ppr,xiao2020visual,yang2018shuffle} where a model is trained to use triplet annotation available at the image level. Different than us, Peyre \etal \cite{peyre2017weakly} leverage a fixed vocabulary of pre-trained object detectors and learn relations with a discriminative clustering model.
% while Zhang \etal \cite{zhang2017ppr} select and classify region pairs obtained by object proposals. 
% However, both approaches model fixed vocabulary and not open vocabulary. 
% More recently, 
Peyre \etal \cite{Peyre19} model open language but in the strongly supervised setting and for still images.

\textbf{Human-object interaction detection.}
Human-object interaction detection \cite{chao2015hico,wang2020discovering,wang2019deep,qi2018learning,zhou2019relation,wan2019pose} is a kind of human-centric relation detection. HOI is an essential research topic for deeper scene understanding.
Several datasets, such as HICO-DET \cite{chao2018learning} and V-COCO \cite{gupta2015visual}, have been proposed for this domain.
% \cite{chao2018learning} propose a multi-stream model combining visual features and spatial location features to detected humans, objects, and human-object spatial relations.
\cite{shen2018scaling,xu2019learning,hou2020visual} formulate the novel HOI detection as a zero-shot learning problem.
However, these methods are based on still images and have difficulties in detecting dynamic human-object interactions.
% that require the scene understanding over time. 
They either rely on the bounding box annotations or pretrained object detectors which has been show perform badly in videos \cite{fouhey2018lifestyle}.

% The main contribution of this work is a model for learning to detect human-object interactions from videos in the weakly supervised setting with open language.

% Nonetheless, we are the first to look at learning to detect visual relations (person-object interactions) from video in the weakly supervised setting with open language.

% \bryan{Pull relation detection references from this paragraph and put above.}
% Research on visual relationship detection \cite{krishna2018referring, jae2018tensorize, yang2018shuffle, yang2018visual, zhang2019large} has received increasing attention in recent years. 
% Lu \etal \cite{lu2016visual} uses language priors from word embeddings to assist the relationship prediction. \cite{zhang2019large} laern a visual and a semantic module to map the visual and textual features to a shared space. Dai \etal \cite{dai2017detecting} propose a deep relational network for exploiting the statistical dependencies between objects and their relationships. 
% Human-object interaction \cite{ramanathan2015learning,yao2010grouplet,yao2010modeling,yao2011human,lu2016visual,galleguillos2008object,johnson2015image,li2012automatic,elhoseiny2017sherlock,choi2012context, gao2018ican, bansal2019detecting, li2019transferable, chao2018learning}. 
% \cite{prest2011weakly} introduces a weakly supervised approach for learning human object interactions. Their model learns to localize a human in the image and then determine the object relevant for a given annotated action.

\vspace{-3pt}
\section{Learning contrastive spatiotemporal regions}
\begin{figure}[t]
\begin{center}
\includegraphics[width=0.5\textwidth]{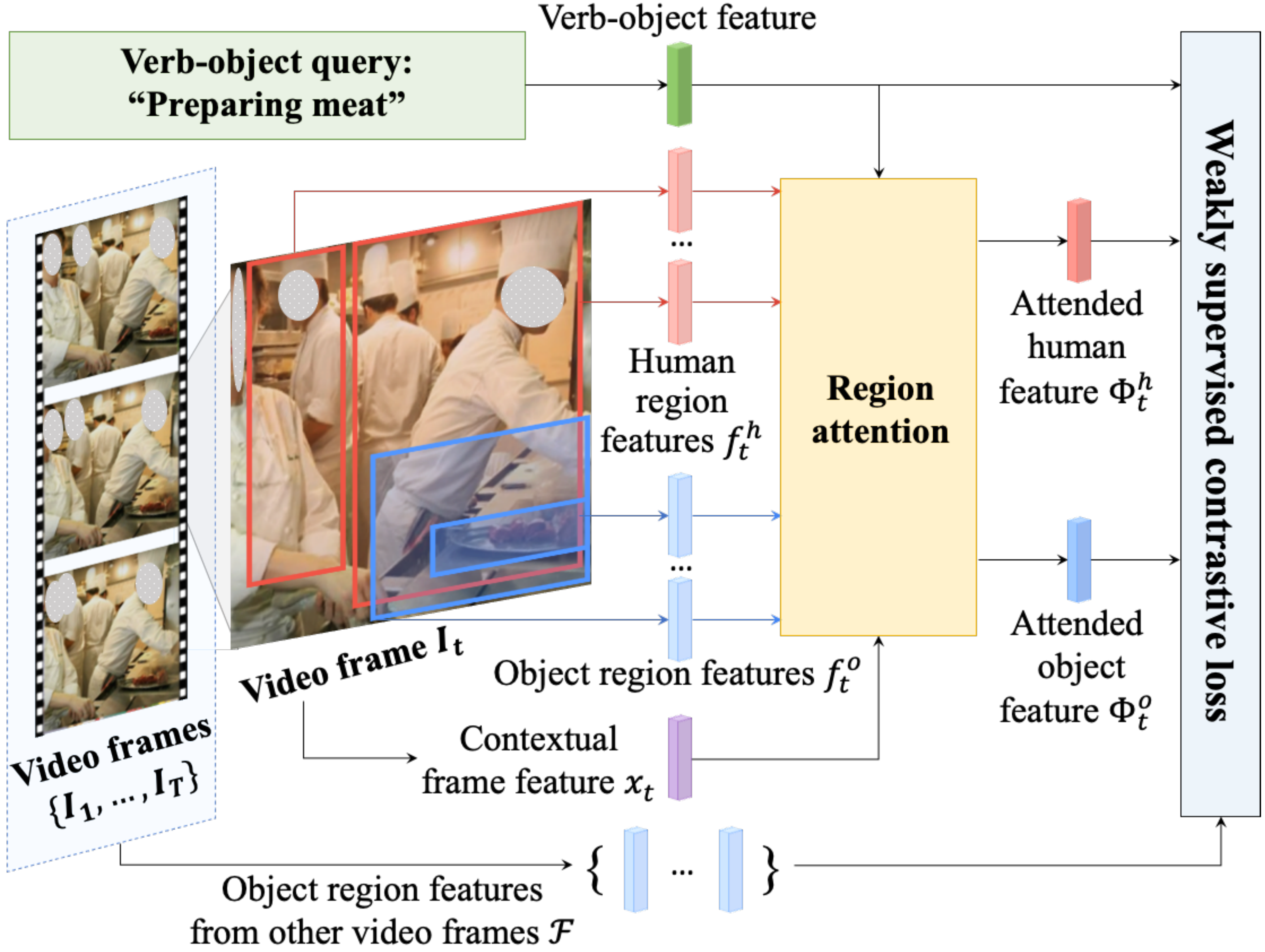}
\end{center}
\vspace{-17pt}
\caption{ \small{\textbf{Training overview.} Given a video clip and a verb-object query, for each frame, we first extract its human and object region features. The human/object features are aggregated in a region attention module to attend to regions that are more relevant to the query. The attended human feature, attended object feature, the feature of verb-object query, and object region features from other frames are used to compute our weakly supervised contrastive loss. (Video credit: 
The Best Gallery Craft~\cite{urlfig2})
}}
\label{fig:training_overview}
\vspace{-15pt}
\end{figure}

\vspace{-3pt}
%\subsection{Model Overview}
% \label{sect:overview}
We address the problem of detecting human-object interactions (HOIs) in videos in a weakly supervised manner.
% More specifically, during testing, we seek to return the human-object interaction labels and the corresponding human and object bounding boxes in the given video.
As obtaining ground truth bounding boxes for supervised learning is expensive and time consuming, we seek to learn from a collection of videos where only verb-object phrase annotations are provided for the entire video clip during training. 
We thus propose a weakly supervised framework that incorporates both spatial and temporal information to detect HOIs in videos.
% \bryan{Perhaps include some of the motivation text at the beginning of the next subsection here.}
%
The overall training setup is illustrated in \fig{fig:training_overview}.
%As shown in \fig{fig:training_overview},
Given a video clip and a verb-object query, for each frame, we first extract a bank of features. The features include those for the verb-object query, frame, and human/object regions in the clip. This bank of features passes through a region attention module that outputs two features for the frame -- an attended human feature and an attended object feature that focus attention on regions that are more relevant to the verb-object query. These features, along with the verb-object feature and object region features from the other frames are passed into our weakly supervised contrastive loss. 

\vspace{-3pt}
\subsection{Weakly supervised contrastive loss}
Learning from language labels in a weakly supervised manner is challenging as a system must automatically identify and associate video spatiotemporal regions with the provided phrase annotations. 
Moreover, HOIs typically  follow a long-tailed distribution.
% our verb-object queries come from an open vocabulary with a long-tailed distribution. 
Applying the often used classification loss will not suffice as it requires a fixed vocabulary with similar number of samples for each class. 
Furthermore, a classification loss maximizes the probability of the correct class while suppressing all other classes, which does not allow for less common or unseen objects and verbs. 
Finally, words with similar meanings are not explicitly mapped to nearby locations in the feature space with a classification loss.
% , and in fact, may be considered as negatives.

\begin{figure*}
\begin{center}
\includegraphics[width=0.98\linewidth]{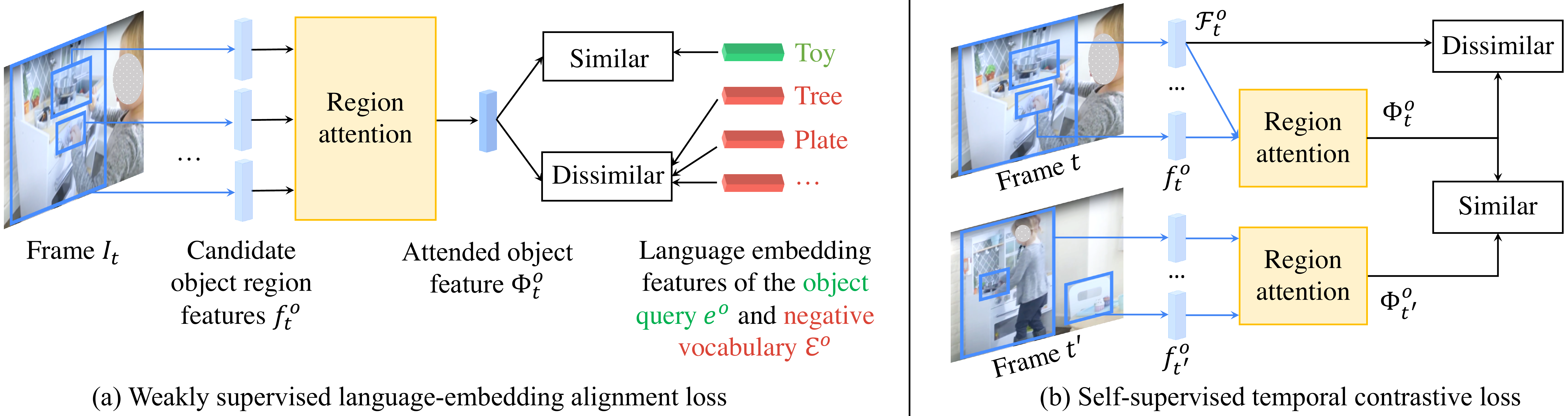}
\end{center}
\vspace{-15pt}
\caption{\small{
{\bf Weakly supervised contrastive loss.} 
	Our loss jointly aligns features for spatiotemporal regions in a video to (a) a language-embedding feature for an input verb-object query and (b) other spatiotemporal regions likely to contain the target object. This figure only shows object regions. The same mechanism is applied to human regions. (Video credit: KidKraft~\cite{urlfig3})
}}
\label{fig:pipeline}
\vspace{-15pt}
\end{figure*}

To address these issues, we introduce a contrastive 
spatiotemporal loss for learning a shared visual-language embedding, as shown in \fig{fig:pipeline}.
Our loss leverages the phrase annotations associated with each training video and cues about the temporal continuity of objects in motion. 
Our training loss incorporates three insights. 
First, we learn to map the visual representation for the likely human and object regions to the corresponding embedded representation of the input verb-object queries and contrast against embedded representations of other non-relevant words in the vocabulary. 
Second, we encourage spatiotemporal regions to be temporally consistent in the video. 
Third, we apply the contrastive loss in our model, enabling it to detect new unseen human-object interactions during testing. 
% an open vocabulary.

% To achieve these goals, we aim to align a learnable visual feature $\bar{\features}$ corresponding to a candidate spatiotemporal region in a video to a positive target feature $\bar{\features}^\prime$ while contrasting against a set of $N$ negative features $\left\{\bar{\features_n}\right\}_{n=1}^N$. 
We build on the contrastive loss~\cite{chen2020simple,henaff2019data,hjelm2018learning}, which aims to encourage
positive pairs of unit-length features to be close (measured by dot product) 
% positive-paired unit-length features to be close 
and negative pairs to be far in the feature space,
\vspace{-7pt}
\begin{equation}
\small
    \losscontrastive{\left( f, f', \{f_n\}_{n=1}^N \right) = -f^T f' + \log\sum_{n=1}^N{\exp(f^T f_n)}},
\label{eqn:contrastive_loss}
\vspace{-3pt}
\end{equation}
where $f$ is an anchor feature, $f'$ is a positive feature and $\left\{f_n\right\}_{n=1}^N$ are $N$ negative features.
We propose a weakly supervised language-embedding alignment loss to align the spatiotemporal regions to the input verb-object query and a self-supervised temporal contrastive loss to encourage temporal continuity of the object regions based on \eqn{eqn:contrastive_loss}.

\textbf{Weakly supervised language-embedding alignment loss.} 
Given a video frame $I_t$, we extract its human and object region proposal features, $f_{t}^h$ and $f_{t}^o$. 
Let $\wordembedding$ be a language-embedding feature for the ground truth verb-object label of the input video. 
We seek to align relevant human/object regions to the ground truth verb-object label. 
Since only the frame-level (or video-level) verb-object label is available, we also seek to learn a global human/object feature in each frame that contrasts against a negative set of language-embedding features $\dictembedding$ covering the vocabulary not including the ground truth verb-object label.

To perform the alignment, we propose a region attention module that computes an attention score $\sigma_{t,i}^h$ and $\sigma_{t,i}^o$ for each human and object region proposal, respectively, to measure their relevance to the verb-object query.
We obtain an attended human feature $\attention^h_t$ by aggregating the human region features $f_{t}^h$ in frame $I_t$ as a weighted average over their attention scores $\sigma_{t}^h$,
\vspace{-7pt}
% \begin{equation}
% \small
%     \attention^h_t = \sum_{i=1}^{|f_t^h|} \frac{\exp(\sigma_{t,i}^h)}{\sum_{j=1}^{|f_t^h|} \exp(\sigma_{t,j}^h)} f_{t,i}^h,
% \label{eqn:feature_aggregation}
% \vspace{-5pt}
% \end{equation}
\begin{equation}
\small
    \attention^h_t = \sum_{i=1}^{N_h} \sigma_{t,i}^h f_{t,i}^h,
\label{eqn:feature_aggregation}
\vspace{-7pt}
\end{equation}
where $N_h$ is the number of candidate human regions. The attended object feature $\attention^o_t$ has a similar form.
The feature attention ``softly'' selects a small number of candidate human/object regions as targets, with higher-scoring regions contributing more to the attended feature.

% \begin{figure*}[t]
% 	\centering
% 	\begin{subfigure}[b]{0.55\linewidth}
% 	\centering
%     \includegraphics[width=\linewidth]{fig/lang_loss.pdf}
%     \caption{\small Language-embedding alignment loss \label{fig:language}}
%     \end{subfigure}
%     \vrule
% 	\begin{subfigure}[b]{0.41\linewidth}
%     \includegraphics[width=\linewidth]{fig/temporal_loss.pdf}
% % 	\vspace{0.02\linewidth}
%     \caption{\small Temporal continuity loss \label{fig:temporal}}
%     \end{subfigure}
%     \vspace{-5pt}
% 	\caption{
% 	{\small {\bf Weakly supervised spatiotemporal loss during training.} 
% 	Our loss jointly aligns features for spatiotemporal regions in a video to (a) a language-embedding feature for an input query and (b) other spatiotemporal regions likely to contain the object. This is only show object, the same thing is applied to human.
% 	}}
% 	\label{fig:pipeline}
% 	\vspace{-10pt}
% \end{figure*}

We define the language-embedding alignment loss $\losslanguage$ as the alignment of the attended features in a frame to the target label while contrasting against the verb or object negative feature set. Following the general expression of the contrastive loss in \eqn{eqn:contrastive_loss}, we define the language-embedding alignment loss in frame $I_t$ as a summation of contrastive losses given attended human/object, language, and negative features,
% \begin{equation}
% \resizebox{1\hsize}{!}{$
%     \losslanguage = \losscontrastive{\left( \Phi\left(f_t^{(h)},\sigma_t^{(h)}\right), \wordembedding^{(h)}, \dictembedding^{(h)} \right)} +
%     \losscontrastive{\left( \Phi\left(f_t^{(o)},\sigma_t^{(o)}\right), \wordembedding^{(o)}, \dictembedding^{(o)} \right)},
% $}
% \end{equation}
\vspace{-5pt}
\begin{equation}
    \losslanguage =
    \losscontrastive{( \Phi^h_t, \wordembedding^v, \dictembedding^v )}  \\
    +
    \losscontrastive{( \Phi^o_t, \wordembedding^o, \dictembedding^o )},
\vspace{-3pt}
\end{equation}
where $e^v$ and $e^o$ are the target verb and object features, respectively, and $\dictembedding^v$ and $\dictembedding^o$ are the negative verb and negative object feature sets, respectively.
More specifically, we rewrite the object term as in \eqn{eqn:contrastive_loss}:
$\small{
\losscontrastive{\left( \Phi^o_t, \wordembedding^o, \dictembedding^o \right)} = 
- (\Phi^o_t)^T \wordembedding^o } + $
$\small{ \log\sum_{n=1}^{N_l}{\exp \left( (\Phi^o_t)^T \dictembedding_n^o \right)}
}$, 
% \begin{equation}
% \resizebox{1\hsize}{!}{$
% \begin{split}
%     &\losscontrastive{\left( \Phi\left(f_t^{(o)},\sigma_t^{(o)}\right), \wordembedding^{(o)}, \dictembedding^{(o)} \right)}  = \\
%     & - \Phi\left(f_t^{(o)},\sigma_t^{(o)}\right)^T \wordembedding^{(o)} + \log\sum_{n=1}^{N_{\text{Lan}}}{\exp \left( \Phi\left(f_t^{(o)},\sigma_t^{(o)}\right)^T \dictembedding_n^{(o)} \right)}, 
% \end{split}
% $}
% \label{eqn:language_loss_human}
% \end{equation}
where $\Phi^o_t$ is the attended object feature that has a similar form as the attended human feature shown in \eqn{eqn:feature_aggregation},
$e^o$ is the target object feature, 
and $N_{l}$ is the number of negative samples in the negative feature set $\dictembedding^o$.
The human term has a similar form.
We show this loss (object term only) in Figure~\ref{fig:pipeline}~(a). 
The “Region attention” module outputs a single “Attended human/object feature” for the video frame. This “Attended human/object feature” forms the positive pair with the verb/object phrase in the corresponding language annotation for the frame. 

% The target human/object region is selected via a soft-attention mechanism (“Region attention” in \fig{fig:pipeline}~(a), which outputs a single “Attended human/object feature” for the video frame. This “Attended human/object feature” forms the positive pair with the verb/object phrase in the corresponding language annotation for the frame. 

\textbf{Self-supervised temporal contrastive loss.} 
We seek to encourage temporal continuity of the moving objects.
We also seek to contrast our learned object features against a negative set of visual features corresponding to likely regions for which the target object does not appear. 
Let $f_{t'}^o$ be a set of features for another frame from the same video with attention scores $\hat{\sigma}_{t'}^o$. 
We define the temporal contrastive loss $\losstemporal$ as the alignment of the attended object feature $\Phi^o_t$ in a frame $I_t$ to the target attended object feature $\Phi^o_{t'}$ in another frame while contrasting against the negative feature set $\featuresnegative_t^o$ from frame $I_t$. Following the contrastive loss in \eqn{eqn:contrastive_loss}, we define the temporal contrastive loss as:
\vspace{-5pt}
\begin{equation}
    \losstemporal = \losscontrastive{\left( \Phi^o_t, \Phi^o_{t'}, \featuresnegative_t^o \right)}.
\vspace{-5pt}
\end{equation}
Note that the attention scores $\hat{\sigma}$ here are different from the soft attention scores $\sigma$ used for the language-embedding alignment loss. In the temporal contrastive loss, we let $\hat{\sigma}$ be hard attention scores, where only one object region has a score of one while the rest of the regions in the same frame has a score of zero. 
In practice, we let the object region that has the highest soft attention score have a hard attention score $\hat{\sigma}=1$, which is the most likely target object described in the verb-object query.
For the negative feature set $\featuresnegative_t^o$, we randomly select from the remaining object regions in frame $I_t$ that are not selected by the hard attention. 
The intuition is that the selected target objects with the highest score from different frames should move consistently through time but should be different from other objects in the same frame.
% We aim at encouraging the temporal continuity of the object that is most likely to be the target object described in the verb-object query.
We illustrate this loss in Figure~\ref{fig:pipeline}~(b).

\textbf{Full weakly supervised contrastive loss.}
We define the final loss at each frame as the sum of the language-embedding alignment and temporal contrastive losses,
\vspace{-5pt}
\begin{equation}
    \lossspatiotemporal = \losslanguage + \alpha \losstemporal,
\label{eqn:spatiotemporalloss}
\vspace{-5pt}
\end{equation}
where $\alpha$ is a hyperparameter. %scalar
Our loss is minimized when a feature corresponding to a softly selected human/object region $\attention_t$ aligns with the language-embedding feature $\wordembedding$ and a similar spatiotemporal region $\attention_{t'}$ in another frame.

% \input{fig_text/attention_model}

% \input{fig_text/human_mask}

%%%%%%%%%%%%%%%%%%%%%%%%%%%%%%%%%%%%%%%%%%%%%%%%%%%%%%%%%%%%%%%%%
%%%%%%%%%%%%%%%%%%%%%%%%%%%%%%%%%%%%%%%%%%%%%%%%%%%%%%%%%%%%%%%%%
%%%%%%%%%%%%%%%%%%%%%%%%%%%%%%%%%%%%%%%%%%%%%%%%%%%%%%%%%%%%%%%%%
% \subsection{Region attention}
% \label{sect:region_attention}

\vspace{-3pt}
\subsection{Feature learning}
\label{sec:feature_learning}
\vspace{-3pt}
In this section, we briefly introduce the object feature, contextual frame feature, and attended human/object features used in \fig{fig:training_overview}. See \sect{sec:model_details} for more details of different types of features.

\begin{figure}
\begin{center}
\includegraphics[width=0.98\linewidth]{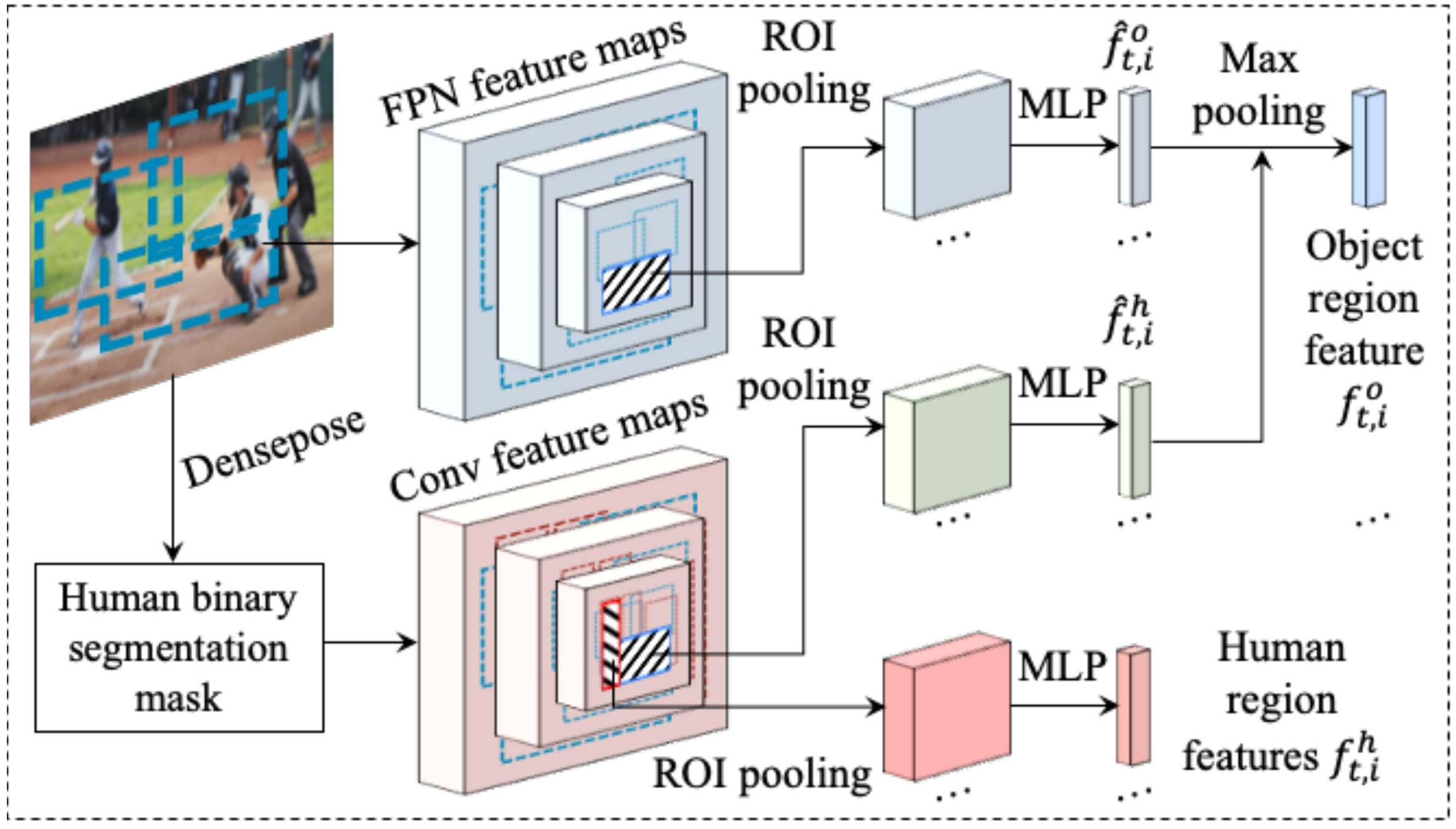}
\end{center}
\vspace{-18pt}
\caption{\small Illustration of extracting human/object features. We learn convolutional filters to encode the Densepose segmentation mask to intermediate features. We obtain the feature of each object region $f_{t,i}^o$ by combining its ROI pooling features from the FPN feature maps $\hat{f}_{t,i}^o$ and the human conv feature maps $\hat{f}_{t,i}^h$. (Video credit: TheOnDeckCircle~\cite{urlfig4})}
\label{fig:human_spatial}
\vspace{-18pt}
\end{figure}

\textbf{Human-guided object feature learning.}
\label{sect:features}
To get the features for candidate object regions, we first extract object location proposals in each video frame using Faster R-CNN \cite{ren2015faster}.
We apply ROI pooling over all the layers of the Faster R-CNN feature pyramid network (FPN) to extract the feature descriptors for the object region proposals.
Each object region proposal has a feature descriptor $\hat{f}^o_{t,i}$ and a bounding box $b^o_{t,i}$ as shown in \fig{fig:human_spatial}.

Human-object interaction is highly correlated with both the human and object features. 
We assume that the spatial co-occurrence of the human and object regions helps to disambiguate the interacted object. 
To more effectively encode the human-object interaction, we incorporate knowledge from the human segmentation masks produced by DensePose \cite{alp2018densepose} into the object proposal features.  
We use ROI pooling to extract a feature $\hat{f}^h_{t,i}$ from the human segmentation mask given the object proposal bounding box $b^o_{t,i}$. 
We apply a max-pool operation over the object region features from the FPN feature maps and the human feature maps to obtain the final object proposal feature $f^o_{t,i}=\max(\hat{f}^o_{t,i}, \hat{f}^h_{t,i})$.

\textbf{Contextual frame feature learning.}
Human-object interactions are temporal events and occur over a period of time. 
To utilize the temporal information from the whole video, we use a soft attention module \cite{vaswani2017attention} to learn a contextual feature representation $x_t$ for each frame.
Given a frame feature $\hat{x}_t$ obtained by passing this frame through a small network, we send $\hat{x}_t$ to an embedding layer to generate a ``query'' feature vector $x^{que}_t$. For the features of all frames $\{x_1, \cdots, x_T\}$ in the same video, we use two different embedding layers to get ``key'' $x^{key}_{t'}$ and ``value'' $x^{val}_{t'}$ vectors.
We compute the inner product of the ``query'' and ``key'' to get a similarity score $s_{t,t'}=(x^{que}_t)^T x^{key}_{t'}$ of the current frame and each frame in the same video. A softmax layer is then applied to the similarity scores to normalize the similarity of each frame to the current frame. 
The contextual frame feature is obtained by the weighted average over frame ``value'' features $x_t = \sum_{t'=1}^T {s_{t,t'} x^{val}_{t'}}$. 
% (See the Supplement for more details.)

\textbf{Region attended human/object feature learning.}
The region attention module computes attention scores for the human/object region proposals to measure their relative relevance to the given verb-object query (\fig{apx_fig:attention_model} in the appendix). 
For each human region in frame $I_t$, we first concatenate its feature representation $f_{t,i}^h$ with the contextual frame feature $x_t$ and the verb-object query feature and then pass them through a small network to obtain a score. We apply the softmax function over the scores of all human regions in this frame and get the final human attention scores $\sigma^h_t$. 
Similarly, each object region has an object attention score $\sigma^o_t$ after applying the softmax function over all object regions.
The attention scores are used to aggregate human/object features using \eqn{eqn:feature_aggregation}.

% For the human attention scores $\sigma_t^h$ at time $t$, we apply the softmax function over all scores and get $\tilde{\sigma}_t^h$ (similarly, we define $\tilde{\sigma}_t^o$ as the scores after the softmax function over all object attention scores). 

% \input{fig_text/attention_model}
% \input{fig_text/testing_overview}

%%%%%%%%%%%%%%%%%%%%%%%%%%%%%%%%%%%%%%%%%%%%%%%%%%%%%%%%%%%%%%%%%
%%%%%%%%%%%%%%%%%%%%%%%%%%%%%%%%%%%%%%%%%%%%%%%%%%%%%%%%%%%%%%%%%
%%%%%%%%%%%%%%%%%%%%%%%%%%%%%%%%%%%%%%%%%%%%%%%%%%%%%%%%%%%%%%%%%
\vspace{-3pt}
\subsection{Training objective}
\label{sect:learning}
\vspace{-3pt}
In addition to the weakly supervised contrastive loss $\lossspatiotemporal$, we propose a sparsity loss $\losssparsity$, and a classification loss $\lossbinary$ for weakly supervised learning.
Our final training loss for a pair of frames is the sum of all the losses,
\vspace{-7pt}
\begin{equation}
    \mathcal{L}_\params(t,t^\prime) = \lossspatiotemporal + \losssparsity + \lossbinary.
\vspace{-5pt}
\end{equation}
We describe the sparsity and classification losses next.

\textbf{Sparsity loss.}
As there are often few humans and objects undergoing the action and object given in the input query, we seek to encourage the attention scores for the human and object proposals each to be high for a single proposal instance and low for all other proposals in each frame. 
To enable this effect, we introduce a sparsity loss which is defined as the sum of negative log $L_2$ norms of the human and object attention scores:
% \begin{equation}
%     \losssparsity = -\log{\left(|\tilde{\sigma}_t^h|_2\right)}-\log{\left(|\tilde{\sigma}_t^o|_2\right)}
% \vspace{-5pt}
% \end{equation}
\vspace{-7pt}
\begin{equation}
    \losssparsity = -\log{\left(|\sigma_t^h|_2\right)}-\log{\left(|\sigma_t^o|_2\right)}
\vspace{-3pt}
\end{equation}
% Since the sums $\sum_i \scoreshumansoftmax_i = \sum_i \scoresobjectsoftmax_i = 1$, the sparsity loss is minimised when one of the proposal scores approaches one while the others approach zero. 

\vspace{-3pt}
\textbf{Classification loss.}
The weakly supervised contrastive loss and sparsity loss enable our model to localize objects and humans given the verb-object query.
To make our model retrieve and localize the language input across videos, we add a classification loss to predict whether the current video contains the interaction described in the verb-object query.
% Even though language labels are much easier to be collected than bounding box annotations, we still want our model can retrieval and localize the mentioned human-object interaction across videos where the ground truth language label of each video is unknown.
% To do this, we add a classification loss to predict whether the current video contain the interaction described in the language label.
In the training phase, each video has a ground truth verb-object label and we assign them a label of $y=1$. We randomly select a negative verb-object label from the language features for the entire vocabulary $\dictembedding$ and assign a label of $y=0$ to the video and the selected negative verb-object label. The classification loss at frame $I_t$ is:
\vspace{-5pt}
\begin{equation}
    \lossbinary = -(y_t \text{log} (p_t^q) + (1-y_t) \text{log} (1-p_t^q) ),
\label{eqn:binary_loss}
\vspace{-3pt}
\end{equation}
where $p_t^q=p(y_t| q, x_t)$ is the likelihood of the input video frame $I_t$ containing the verb-object query $q$. Here $x_t$ is the contextual frame feature of frame $I_t$.

\vspace{-3pt}
\subsection{Inference}
\label{sect:inference}
\vspace{-3pt}
During inference, given a video frame $I_t$, we randomly select one verb-object query $q$ and compute their binary classification score $p_t^q$ as shown in \eqn{eqn:binary_loss}.
Since we encourage the matching pairs of video frame and verb-object query to have higher probability during training, score $p_t^q$ is able to evaluate the probability of the verb-object query appearing in the given frame during inference.
We also have an attention score $\sigma_{t,i}^h$ or $\sigma_{t,j}^o$ for each human or object region proposal, representing their relevance to the given verb-object query.
Thus for each human-object pair, we compute their confidence score as $c_{t,i,j}^q =p_t^q \times (\sigma_{t,i}^h + \sigma_{t,j}^o)/{2}$.
For HOI detection, we predict human and object bounding boxes and their HOI label.
For each video frame, we feed in all possible verb-object labels appearing in the dataset and select the verb-object label having the highest confidence score as the HOI label prediction result for each pair of human and object regions.
% given this pair of human and object regions.
% thus each pair of human and object region proposals has a confidence score of belonging to each verb-object class. 

\begin{table*}[t]
\caption{\small{Evaluation of each component of the proposed model.  Phrase (Phr) detection refers to correct localization (0.3 IoU) of the union of human and object bounding boxes while relation (Rel) refers to correct localization (0.3 IoU) of both human and object bounding boxes.}}
\vspace{-18pt}
\label{tbl:ablations}
\begin{center}
\small
\setlength{\tabcolsep}{3pt}
\resizebox{1\textwidth}{!}{
\begin{tabular}{l|cccc|cc|cc|cc}
\toprule
\bf Model & \multicolumn{4}{c|}{\bf mAP (\%)} & \multicolumn{2}{c|}{\bf Recall@1 (\%)} & \multicolumn{2}{c|}{\bf Video One Recall@1 (\%)} & \multicolumn{2}{c}{\bf Video All Recall@1 (\%)}  \\
\midrule
& Phr (ko) & Phr (def) & Rel (ko) & Rel (def) & Phr & Rel & Phr & Rel & Phr & Rel  \\
\midrule
Baseline(add) & 40.59 & 0.45 & 6.95 & 0.11 & 75.98 & 19.00 & 90.30 & 33.22 & 60.36 & 7.07  \\
Baseline(cat) & 41.86 & 0.35 & 11.34 & 0.11 & 75.93 & 19.91 & 88.49 & 35.53 & 61.68 & 5.92 \\
(cat)+Spa & 50.79 & 1.02 & 16.23 & \bf 0.47 & 79.52 & 24.24 & 87.01 & 38.49 & 69.74 & 9.67 \\
(cat)+Spa+Hum & 55.60 & 0.89 & 15.91 & 0.29 & 81.35 & 22.99 & 91.61 & 38.82 & 70.56 & 9.55 \\
% (cat)+Spa+Hum+Tem & 53.73 & 1.33 & 18.23 & 0.51 & 80.72 & 24.02 & 90.46 & 37.66 & 69.74 & 10.69 \\
(cat)+Spa+Hum+Tem & 54.42 & \bf 1.24 & 16.94 & 0.30 & 81.00 & 25.61 & 91.12 & 39.14 & 69.74 & 12.68 \\
(cat)+Spa+Hum+Tem+Con & \bf 55.90 & 0.90 & \bf 18.56 & 0.26 & \bf 84.08 & \bf 30.12 & \bf 91.94 & \bf 44.90 & \bf 75.33 & \bf 15.95 \\
\bottomrule
\end{tabular}
}
\end{center}
\vspace{-17pt}
\end{table*}

\begin{table*}[t]
\caption{\small{Evaluation of performance on \lsdata compared to methods in \cite{peyre2017weakly}, \cite{zhou2018weakly}, \cite{xiao2020visual} and different random baselines. Phrase (Phr) detection refers to correct localization (0.3 IoU) of the union of human and object bounding boxes while relation (Rel) refers to correct localization (0.3 IoU) of both human and object bounding boxes. (ko) and (def) are the known object setting and default setting.}}
\vspace{-18pt}
\label{tbl:other}
\begin{center}
\setlength{\tabcolsep}{4pt}
\resizebox{0.9\textwidth}{!}{
\begin{tabular}{l|cccc|cc|cc|cc}
\toprule
\bf Model & \multicolumn{4}{c|}{\bf mAP (\%)} & \multicolumn{2}{c|}{\bf Recall@1 (\%)} & \multicolumn{2}{c|}{\bf Video One Recall@1 (\%)} & \multicolumn{2}{c}{\bf Video All Recall@1 (\%)}  \\
\midrule
& Phr (ko) & Phr (def) & Rel (ko) & Rel (def) & Phr & Rel & Phr & Rel & Phr & Rel  \\
\midrule
Random & 11.24 & 0.08 & 0.57 & 0.00 & 22.42 & 4.05 & 40.79 & 8.88 & 6.25 & 0.49 \\
Random Pretrain & 9.58 & 0.02 & 0.48 & 0.00 & 12.26 & 3.59 & 25.33 & 8.06 & 1.97 & 0.33 \\
\cite{peyre2017weakly} & 32.42 & 0.14 & 2.06 & 0.01 & 45.75 & 5.02 & 71.38 & 14.14 & 20.72 & 0.16 \\
\cite{zhou2018weakly} & 21.88 & 0.60 & 4.83 & 0.04 & 55.56 & 8.04 & 71.05 & 16.45 & 38.49 & 1.97 \\
\cite{xiao2020visual} & 25.34 & 0.12 & 4.06 & 0.05 & 43.07 & 5.31 & 63.16 & 12.50 & 24.84 & 0.49 \\
\midrule
% Ours & \bf 55.78 & \bf 1.10 & \bf 19.38 & \bf 0.22 & \bf 83.51 & \bf 28.35 & \bf 90.79 & \bf 41.28 & \bf 74.18 & \bf 15.30 \\
% Ours & \bf 55.49 & \bf 1.18 & \bf 19.50 & \bf 0.40 & \bf 83.28 & \bf 27.95 & \bf 90.79 & \bf 41.28 & \bf 74.01 & \bf 15.30 \\
Ours & \bf 55.90 & \bf 0.90 & \bf 18.56 & \bf 0.26 & \bf 84.08 & \bf 30.12 & \bf 91.94 & \bf 44.90 & \bf 75.33 & \bf 15.95 \\
\bottomrule
\end{tabular}
}
\end{center}
\vspace{-22pt}
\end{table*}

\vspace{-7pt}
\section{Human-object interaction video dataset}
\vspace{-5pt}
Existing human-object interaction datasets either focus on classification~\cite{chao2015hico} or detection in static images~\cite{chao2018learning,gupta2015visual}. However, human-object interaction is a temporal process and it is more naturally done in video data.
Current video datasets, such as Charades~\cite{YuanyuanICCV2017}, EpicKitchens~\cite{damen2018scaling}, VidVRD~\cite{shang2017video},  VidOR~\cite{shang2019annotating}, 
% ActivityNet Entity~\cite{caba2015activitynet}, 
and YouCook~\cite{ZhXuCoCVPR18,ZhLoCoBMVC18} are not suitable for human-object interaction detection. 
First, most of them do not have human bounding box annotations. Second, all objects in a scene are annotated, with annotated objects not necessarily interacting with humans. Furthermore, EpicKitchens and YouCook do not have triplet human-action-object labels. VidVRD and VidOR are for visual relation detection and the relations are not necessarily human-centric.
Thus, they cannot be directly used for evaluating video based human-object interaction detection.

Instead, to study the human-object interaction problem in videos, we collect a large, diverse Video dataset of Humans Interacting with Common Objects (\lsdata). Our dataset has a large variety of actions and interacted objects. Our dataset has more videos (6,594) than Epic-Kitchens (432) and YouCook (2,000), with each video containing human-object interactions. Furthermore, the new dataset is more challenging with more diverse outdoor scenes compared with Charades, EpicKitchens, and YouCook that either focus on household or kitchen scenes.

\begin{table*}[t]
\footnotesize
\caption{\small 
Evaluation of our proposed approach, \cite{peyre2017weakly}, and different random baselines on the unseen test set on \lsdata. The unseen test set consists of 51 classes of objects unseen during training. 
Evaluation at IoU threshold 0.3. 
% (ko) and (def) are the known object setting and default setting.
}
\vspace{-17pt}
\label{tbl:unseen}
\begin{center}
\setlength{\tabcolsep}{4pt}
\resizebox{0.92\textwidth}{!}{
\begin{tabular}{l|cccc|cc|cc|cc}
\toprule
\bf Model & \multicolumn{4}{c|}{\bf mAP (\%)} & \multicolumn{2}{c|}{\bf Recall@1 (\%)} & \multicolumn{2}{c|}{\bf Video One Recall@1 (\%)} & \multicolumn{2}{c}{\bf Video All Recall@1 (\%)} \\
% \midrule
\midrule
& Phr (ko) & Phr (def) & Rel (ko) & Rel (def) &  Phr & Rel & Phr & Rel & Phr & Rel \\
\midrule
Random  & 10.44 & 0.16 & 0.74 & 0.03 & 14.79 & 2.11 & 26.92 & 5.77 & 7.69 & 0.00 \\
Random  Pretrain & 4.78 & 0.10 & 0.42 & 0.02 & 14.79 & 2.82 & 28.85 & 5.77 & 1.92 & 0.00 \\
Peyre2017 \cite{peyre2017weakly} & 38.19 & 0.70 & 4.79 & 0.07 & 43.24 & 5.41 & 64.81 & 12.96 & 16.67 & 0.00  \\
\midrule
% Ours & \bf 67.73 & \bf 2.26 & \bf 23.43 & \bf 0.69 & \bf 90.28 & \bf 31.94 & \bf 96.15 & \bf 42.31 & \bf 78.85 & \bf 19.23 \\
Ours & \bf 67.21 & \bf 2.76 & \bf 25.10 & \bf 0.66 & \bf 91.89 & \bf 31.08 & \bf 94.44 & \bf 42.59 & \bf 85.19 & \bf 18.52 \\
\bottomrule
\end{tabular}
}
\end{center}
\vspace{-25pt}
\end{table*}

Our \lsdata dataset contains 5,297 training videos, 635 validation videos, 608 test videos, and 54 unseen test videos of human-object interactions. 
To test the performance of models on common human-object interaction classes and generalization to new human-object interaction classes, we provide two test splits, the first one has the same human-object interaction classes in the training split while the second one consists of unseen novel classes. 
Our training set consists of 193 object classes and 94 action classes. There are 653 action-object pair classes in the training set.
The unseen test set contains 51 object classes and 32 action classes with 52 action-object pair classes.
All videos are labeled with text annotations of the human action and the associated object. 
The test set and the unseen test set contain the annotations of both human and object bounding boxes. 

Our `unseen' test set (51 unseen object classes) contains 2 classes present in the MSCOCO object vocabulary, 8 present in OpenImages, and 34 present in VisualGenome.
We use the object detector pre-trained on MSCOCO, indicating only 2 object classes have been seen during pre-training.
Furthermore, our entire dataset has 244 object classes in total.
156 of them are not present in MSCOCO or OpenImages,
\eg, 'javelin', and hence cannot be detected using detectors
pre-trained on those datasets. The object distribution
is long-tailed and many objects do not have annotated training data in the publicly available object datasets.
Our model provides a way to scale-up to a large
set of objects without relying on bounding box annotations.

\vspace{-3pt}
\section{Experiments}
\vspace{-3pt}
We evaluate the ability of our method and baselines to detect human-object interactions on the \lsdata dataset.
% and HICO-DET \cite{chao2018learning} image dataset. 

% We first introduce the protocols for evaluation in \sect{sect:metrics}. Then we perform the ablation studies of each model component in \sect{sect:ablation} and compare the proposed method with other baselines on the test and unseen datasets in \sect{sect:baselines} and \sect{sect:unseen}.
% We also present some qualitative result in \sect{sect:qualitative}.

% \subsection{Experiments on \lsdata}
% \label{sect:exp-vhico}
\vspace{-4pt}
\subsection{Evaluation criteria}
\vspace{-4pt}
\label{sect:metrics}
We evaluate the proposed method and other approaches under two settings -- phrase accuracy and relation accuracy.   
%We use two setups, \ie, \textbf{phrase} and \textbf{relation}, to evaluate the proposed model and baselines. 
We denote \textbf{phrase} accuracy when the union of the detected human and object bounding boxes matches the union of the ground truth human and object boxes. 
We denote \textbf{relation} accuracy when both the predicted human and object bounding boxes match the ground truth human and object bounding boxes, respectively. 
Relation accuracy is lower than phrase accuracy since it is more strict on the predicted human and object bounding boxes.
% Correctly localizing the  is harder than the phrase.

We report the mean average precision (mAP) and Recall in these two setups. 
For mAP, we follow the settings proposed by HICO-DET \cite{chao2018learning}. They proposed two different evaluation settings: (1) Known Object setting (\textbf{ko}): Given a human-object interaction category, they evaluate the human and object detection only on images containing the target object category. Here we use video frames that contain the target HOI category. (2) Default setting (\textbf{def}): Given a HOI category, they evaluate the detection on the full test set. This setting is more challenging because it requires models to distinguish whether an image/frame contains the target HOI category and to localize the target HOI simultaneously. 
Note that the evaluation metric we used is designed for HOI detection \cite{chao2018learning}, which is a harder problem than language grounding.
In language grounding, the query input appears in the video and the models return its corresponding bounding box during test. However, in the \textbf{def} setting, the query input does not necessarily appear in the video.
% The models need to predict whether a video has the given query and human/object bounding boxes simultaneously. 
% The outputs are bounding boxes with labels which is a detection problem.

For each frame, we extract the top 10 predicted pairs of human-object bounding boxes based on their score $c_{t,i,j}^q$ as described in \sect{sect:inference}.
The predicted human and object bounding boxes are treated as correct if their Intersection-over-Union (IoU) with ground truth human and object bounding boxes is larger than 0.3 for both the phrase and relation accuracy, similar to~\cite{peyre2017weakly}. We follow HICO-DET \cite{chao2018learning} and compute the \textbf{mAP} over all verb-object classes.
% We construct a precision recall curve to compute the average precision (AP) over all annotated video frames in each verb-object class and average the APs across all classes. 

We also report the frame recall of the top-1 prediction. 
Given a frame and its true verb-object label, we test if the top-1 predicted human-object bounding-box pair matches the ground truth bounding boxes. 
\textbf{Recall@1} is the number of frames where the predictions are correct divided by the number of all frames.
% Based on our dataset, 
We also propose two video recall settings.
In \textbf{Video One Recall}, if all of the ground truth human-object pairs in one frame are detected, the video is considered correct.
% the video is treated as correct if the ground truth human-object pairs are detected in any of the frames (the top-1 prediction is correct). 
Video One Recall is the number of correct videos divided by the number of all videos. In \textbf{Video All Recall}, the video is correct only when all of the ground truth human-object pairs in all frames are detected.

\begin{figure*}[h]
\begin{center}
% height=0.5\columnwidth, 
\includegraphics[ width=1\linewidth]{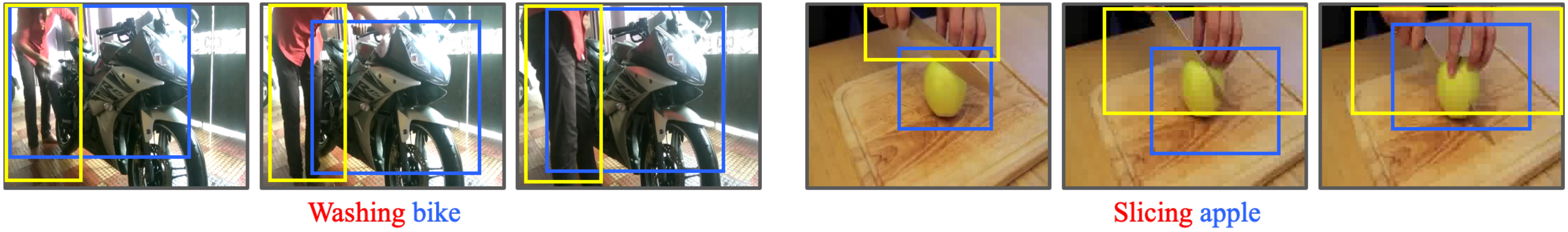}
\end{center}
\vspace{-20pt}
\caption{\small Qualitative predictions of our model with top predicted human bounding boxes (yellow) and object bounding boxes (blue). (Video credit: Dude Chennai~\cite{urlfig1} and Serious Eats~\cite{urlfig5})}
\label{fig:qual_lsdata}
\vspace{-20pt}
\end{figure*}

% \begin{figure*}[t]
% \begin{center}
% % height=0.5\columnwidth, 
% \includegraphics[ width=1\linewidth]{fig/results2.pdf}
% \end{center}
% \vspace{-15pt}
% \caption{\small Qualitative prediction of our model on the \lsdata test set with predicted human bounding box (red), predicted object bounding box (green), and input action-object label (bottom).}
% \label{fig:qual_lsdata}
% \vspace{-15pt}
% \end{figure*}

\vspace{-3pt}
\subsection{Ablation studies on \lsdata}
\vspace{-3pt}
\label{sect:ablation}
% \bryan{This subsubsection needs to be updated to align with the new text for the approach.} 
% \shuang{will update}
To investigate the effect of each component of our approach, we perform a series of ablation studies on our \lsdata dataset. 
We report the results in \tbl{tbl:ablations}.
We first evaluate our approach when no temporal continuity is enforced in the model during training. To achieve this goal, we omit the temporal contrastive loss $\losstemporal$ and the sparsity loss $\losssparsity$ during training and do not include the human ROI-pooled feature as part of the object proposal feature.
We investigate different ways to merge the human/object region features, verb-object language features, and the frame feature $\hat{x}_t$ (before the temporal soft attention described in \sect{sec:feature_learning}) when computing human/object attention scores.
We find that feature addition (\textbf{Baseline(add)}) and feature concatenation (\textbf{Baseline(cat)}) have similar results. 
% Feature concatenation yields slightly better accuracy overall. 

Next, we evaluate the efficacy of the sparsity loss $\losssparsity$.
Without the sparsity loss, empirically we find that the output attention scores are often uniformly distributed across all region proposals.
%To encourage our model to focus on few human and object candidates, we add a sparsity loss $\losssparsity$.
\textbf{(cat)+Spa} is the result after using the sparsity loss based on the feature concatenation baseline; it boosts the performance significantly.
% We observe that the sparsity loss (\textbf{(cat)+Spa}) boosts our performance on recalls. 

As existing human detectors are quite robust in videos and the spatial location of the human can help localize the interacted object, we evaluate the effect of including the ROI-pooled feature from the human segmentation feature maps to the object region feature (\textbf{(cat)+Spa+Hum}). 
We observe that including the human information when learning object features improves the performance.
Next, we evaluate the efficacy of including our self-supervised temporal  contrastive loss $\losstemporal$.
\textbf{(cat)+Spa+Hum+Tem} improves the performance by encouraging the temporal continuity of moving objects.
We investigate the effect of the contextual frame feature generated using the soft attention over the entire video. \textbf{(cat)+Spa+Hum+Tem+Con} is the result of adding the contextual frame feature and is used as the final result of the proposed model.

% We investigate the effect of contextual frame feature generated using the soft attention over the entire video.
% \textbf{(cat)+Spa+Hum+Tem+Con} is the final result of the proposed model after adding the contextual frame feature.

To further verify the contribution of human information to object detection results, 
we add a baseline that localizes the object based on the human spatial prior alone.
We first use our model to generate candidate human/object proposals and their confidence scores.
We select the human proposal with the highest score as the target human.
For each object proposal, we compute its confidence as the inverse distance of its centroid to the human proposal.
The mAP on Phr (ko) is $46.67$ while ours is $55.90$. 
Note that this baseline's mAP is not too bad as it uses human/object proposals and human confidence scores from our trained model (\textbf{(cat)+Spa+Hum+Tem+Con}),
yet it performs significantly worse than our full model.

\vspace{-4pt}
\subsection{Comparison with baselines}
\vspace{-4pt}
\label{sect:baselines}
Since most existing HOI approaches use supervised learning on static images, we compare with the three most related methods \cite{peyre2017weakly,zhou2018weakly,xiao2020visual} and add two random baselines to compare with our approach in \tbl{tbl:other}. 
% We compare our proposed approach with five methods on \lsdata in \tbl{tbl:other}. 
% All the baselines use the same input, \ie, action and object labels, as the proposed model.
``Random'' is our model using randomly initialized parameters.
``Random Pretrain'' is our model with the Faster R-CNN part initialized from the COCO pretrained model and other parts randomly initialized as above.

Since there is no existing weakly supervised human-object interaction detection method for videos, we modify three related weakly supervised baselines using their publicly available code. 
Peyre \etal~\cite{peyre2017weakly} is a weakly supervised approach for visual relation detection in single still images. 
For fair comparison, our implementation of Peyre \etal~\cite{peyre2017weakly} uses the same human and object bounding boxes and features generated by DensePose and Faster R-CNN, respectively, as our approach. 
For each human-object bounding box pair, the classifier predicts its probability score of being each human-object interaction class. 
% We extract the confidence score of the ground truth class using object and action labels. 
The human-object bounding box pairs are ranked based on their confidence scores for evaluation.
We also compare with Zhou \etal~\cite{zhou2018weakly}, a method for video-based object grounding from text. We use the same Faster R-CNN to generate object bounding box proposals. A human detection branch is added using the human proposals generated by DensePose.
We further modified a video relation grounding method \cite{xiao2020visual} which also uses the Densepose and Faster R-CNN to generate human and object bounding boxes for a fair comparison.
% human and object bounding boxes and features generated Densepose and Faster R-CNN, language ranking scores for action are only applied over Densepose bounding boxes while ranking scores for object is applied over Faster R-CNN bounding boxes. 

\tbl{tbl:other} shows the comparisons of our approach and these baseline methods on our \lsdata dataset.
Overall, our model outperforms all the baselines as
\cite{peyre2017weakly} is an image-based method without taking advantage of the video information, 
\cite{zhou2018weakly} optimizes object and human bounding boxes and features separately without explicitly considering human-object interactions, and
\cite{xiao2020visual} uses a spatiotemporal region graph that may accumulate errors over time.

\vspace{-3pt}
\subsection{Comparison with baselines on unseen classes}
\label{sect:unseen}
\vspace{-3pt}
To test the generalization ability of our model on unseen objects, we evaluate our method on 52 unseen verb-object classes from the unseen dataset.
% \bryan{I don't quite understand the next sentence.} 
Note that there are difficulties when evaluating the Zhou \etal \cite{zhou2018weakly} and Xiao \etal \cite{xiao2020visual} baselines on the unseen dataset as most object labels do not appear in the training set. 
Zhou \etal \cite{zhou2018weakly} optimize the word embedding for object and action classes during training; for unseen objects and actions, they do not have an optimized word embedding. 
While Xiao \etal \cite{xiao2020visual} report results on zero-shot relation grounding, they consider the case when the subject-predicate-object triplet is never seen but the separate subject, predicate or object are known during training. 
However, on our unseen test set, most object labels do not
appear in the training set.
Thus Xiao \etal \cite{xiao2020visual} have the same problem as Zhou \etal \cite{zhou2018weakly} -- they do not have an optimized embedding for unseen words.
Thus we only compare our method with Peyre \etal \cite{peyre2017weakly}, ``Random'', and ``Random Pretrain'' on the unseen test set.

\tbl{tbl:unseen} shows that our method generalises well to new object classes and significantly outperforms the baselines in terms of both the phrase and relation accuracy.
Our approach on the unseen test set is better than the seen test set as the unseen set is smaller and easier (54 videos, most scenes have a single human) than the seen set (608 videos, more challenging scenes with multiple humans or blurry objects). 
The size of the test set influences some criteria, \eg, mAP is computed over bounding boxes from all videos, thus mAP tends to be lower if there are more videos in the test set. 
% These differences do not allow drawing conclusions about the relative accuracy trends between the two test sets for our method and the baselines.

\vspace{-3pt}
\subsection{Qualitative results}
\label{sect:qualitative}
\vspace{-3pt}
\textbf{Human-object interaction detection results.}
We present the human and object bounding box predictions of our model in \fig{fig:qual_lsdata}. We only show the top 1 human-object bounding box pair.
The yellow bounding boxes represent the predicated human bounding boxes while the blue bounding boxes are the predicated object bounding boxes. We find that the proposed weakly supervised method tends to generate large object bounding boxes as learning from weak supervision is challenging. The system must automatically identify and associate video spatiotemporal regions with the provided phrase annotations during training.

% The first two rows are the results on the test dataset while the thrid row shows the results on the unseen dataset. 
% %
% The last row shows 8 different failure cases of our model. Images 1-3 are due to failure of the Densepose detection. Humans are incorrectly detected in the first two images while missing in the last image. Images 4-5 are because of the ambiguity of target object as they are collocated with human and always co-occur together. Images 6-7 are due to the objects are too small to be detected. Image 8 is because the large number of similar objects are hard to distinguish.

% \input{fig_text/failure}

\textbf{Failure case analysis.}
We notice three main failures in our model predictions: (1) when the human prediction is wrong due to incorrect Densepose output (\eg, missed detections when only a small human body part is visible or when multiple people cause heavy occlusions), (2) when the object prediction is incorrect because the object is small, moving, blurry, or partially occluded, and (3) when both detections are incorrect in challenging scenes, \eg, nighttime.

% Weakly supervised HOI detection in videos is a challenging problem, which  has not yet received much attention.
% However, this problem is of great importance as human-object interactions are common in real life with many important applications, such as autonomous driving, surveillance, and human-robot interaction. 
% Our method is the first step in  this direction, with the failure cases showing that there is still a large space for further improvements.  

\vspace{-5pt}
\section{Conclusion}
\vspace{-5pt}
Weakly supervised HOI detection in videos is a challenging problem, which  has not yet received much attention.
However, this problem is of great importance as human-object interactions are common in real life with important applications, such as video search and editing, surveillance, and human-robot interaction. 
In this paper, we introduce a contrastive loss for learning to detect humans and interacted objects in videos given weak supervision. We demonstrate our approach on a new dataset of videos with verb and object phrase annotations. 
Our approach is a step toward understanding everyday human-object interactions in videos.
We hope the proposed dataset and method can facilitate future research in this direction.

%contrastively trained weakly supervised model for human object relation detection. We train \model on new diverse video dataset for human object relations. We show that our model, combining human and temporal information is able outperform alternative approaches on human object relation detection.

\textbf{Acknowledgments.} 
This work was partly supported by the European Regional Development Fund under the project IMPACT (reg. no. CZ.02.1.01/0.0/0.0/15 003/0000468).

\appendix

\noindent\textbf{\huge{Appendix}}

\vspace{10pt}
In this appendix, we first give the model architecture details and implementation details in \sect{sec:details}. Then we provide the dataset collection details in \sect{sec:dataset_collection_details}. In \sect{sec:dataset_statistics}, we show the dataset statistics.

\section{Model architecture details and implementation details}
\label{sec:details}
We provide the model architecture details in \sect{sec:model_details} and implementation details in \sect{sect:implementation}.

\subsection{Feature learning}
\label{sec:model_details}
In Section 3.2 of the main paper, we introduce the object feature, contextual frame feature, and attended human/object features. In this section, we provide more details about the different types of features.

\textbf{Verb-object query feature learning.}
To extract the feature of the verb-object query described in the main paper (Figure 2), we first map the input verb and object queries to embedded features $\hat{e}^v$ and $\hat{e}^o$, respectively, using the publicly available Google News Word2Vec model \cite{mikolov2013efficient}. 
Next, we pass the embedded features through linear mappings $W_v$ and $W_o$ to obtain 128-dimensional vectors $e^v = W_v \hat{e}^v$ and $e^o = W_o \hat{e}^o$.

\begin{figure}
\begin{center}
\includegraphics[width=0.97\linewidth]{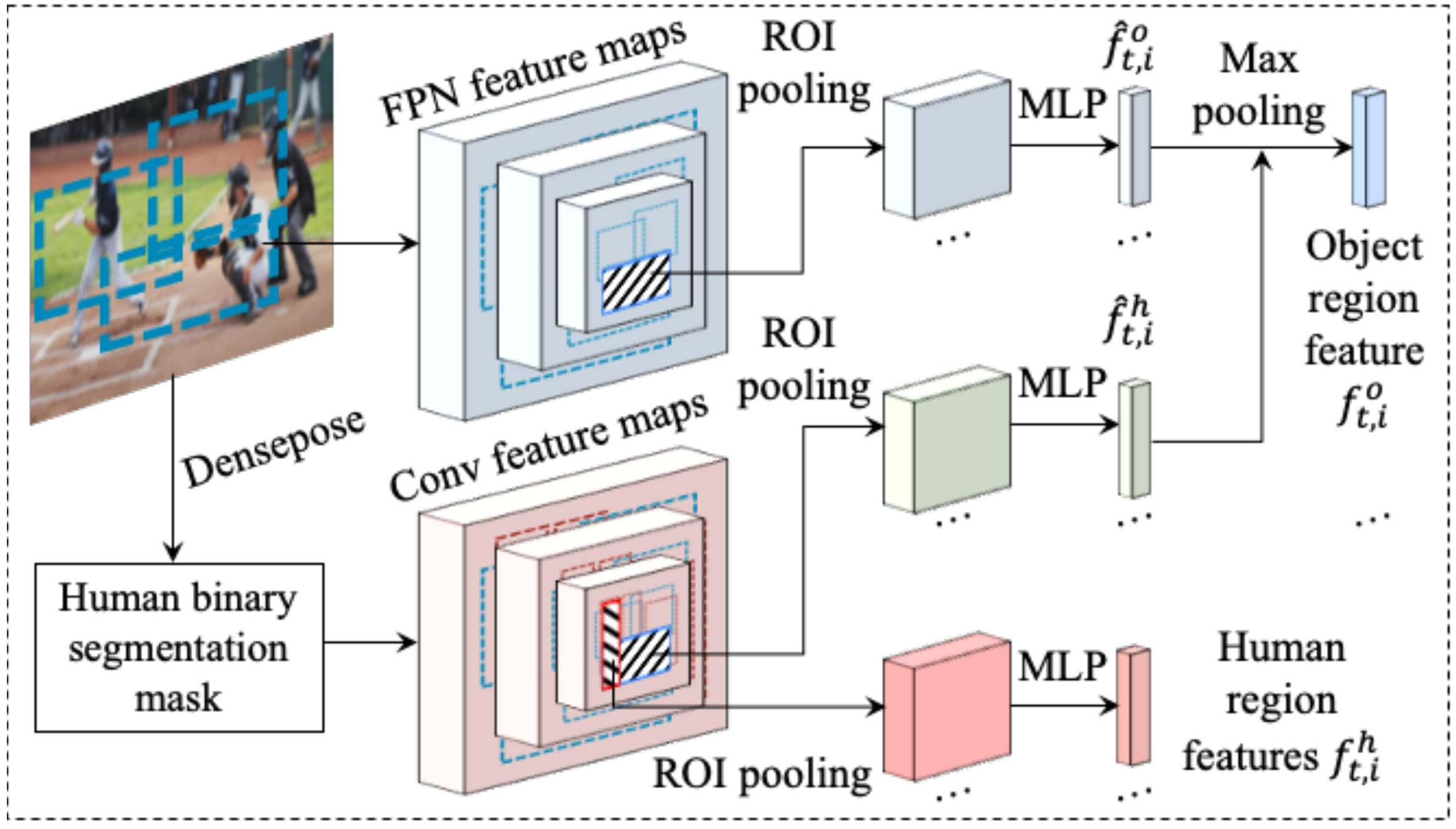}
\end{center}
\vspace{-15pt}
\caption{\small \textbf{Illustration of extracting human/object features.} We learn convolutional filters to encode the Densepose segmentation mask to intermediate features. We obtain the feature of each object region $f_{t,i}^o$ by combining its ROI pooled features from the FPN feature maps $\hat{f}_{t,i}^o$ and the human conv feature maps $\hat{f}_{t,i}^h$. The feature of each human region $f_{t,i}^h$ is the corresponding ROI pooled feature from the human conv feature maps. (Video credit: TheOnDeckCircle~\cite{urlfig4})}
\label{apx_fig:human_spatial}
\vspace{-15pt}
\end{figure}

\textbf{Human feature learning.}
To get the human region features $f_t^h$ described in the main paper (Figure 2), we first extract candidate human location proposals in each video frame using the publicly available DensePose model \cite{alp2018densepose}, which returns a binary segmentation mask of humans in the scene and human bounding-box proposals. 
As shown in \fig{apx_fig:human_spatial}, at time $t$, each human region proposal $i$ has a bounding box $b^h_{t,i}$.
We pass the segmentation mask to a convolutional network to generate human feature maps and then use ROI pooling over the human bounding box $b^h_{t,i}$ to generate human region features $f^h_{t,i}$. 
The convolutional network consists of a $7\times7$ spatial convolutional layer, followed by ReLU and max-pooling nonlinearities, followed by a $3\times3$ spatial convolutional layer.

% \textbf{Human-guided object feature learning.}
% To get the features for candidate object regions, we first extract object location proposals in each video frame using Faster R-CNN \cite{ren2015faster}.
% We apply ROI pooling over all the layers of the Faster R-CNN feature pyramid network (FPN) to extract the feature descriptors for the object region proposals.
% Each object region proposal has a feature descriptor $\hat{f}^o_{t,i}$ and a bounding box $b^o_{t,i}$ as shown in \fig{fig:human_spatial} of this supplement.

% Human-object interaction is highly correlated with both the human and object features. 
% We assume that the spatial co-occurrence of the human and object regions helps to disambiguate the interacted object. 
% To more effectively encode the human-object interaction, we incorporate knowledge from the human segmentation masks produced by DensePose \cite{alp2018densepose} into the object proposal features.  
% We send the human segmentation masks to a convolutional network to generate human feature maps and use ROI pooling over the object proposal bounding box $b^o_{t,i}$ to extract a human-guided object feature $\hat{f}^h_{t,i}$ from the human feature maps (the human feature maps are the same feature maps used in the ``human feature learning'' part). 
% We apply a max-pool operation over the object region features from the FPN feature maps and the human feature maps to obtain the final object proposal feature $f^o_{t,i}=\max(\hat{f}^o_{t,i}, \hat{f}^h_{t,i})$.

\begin{figure}
\begin{center}
\includegraphics[width=1\linewidth]{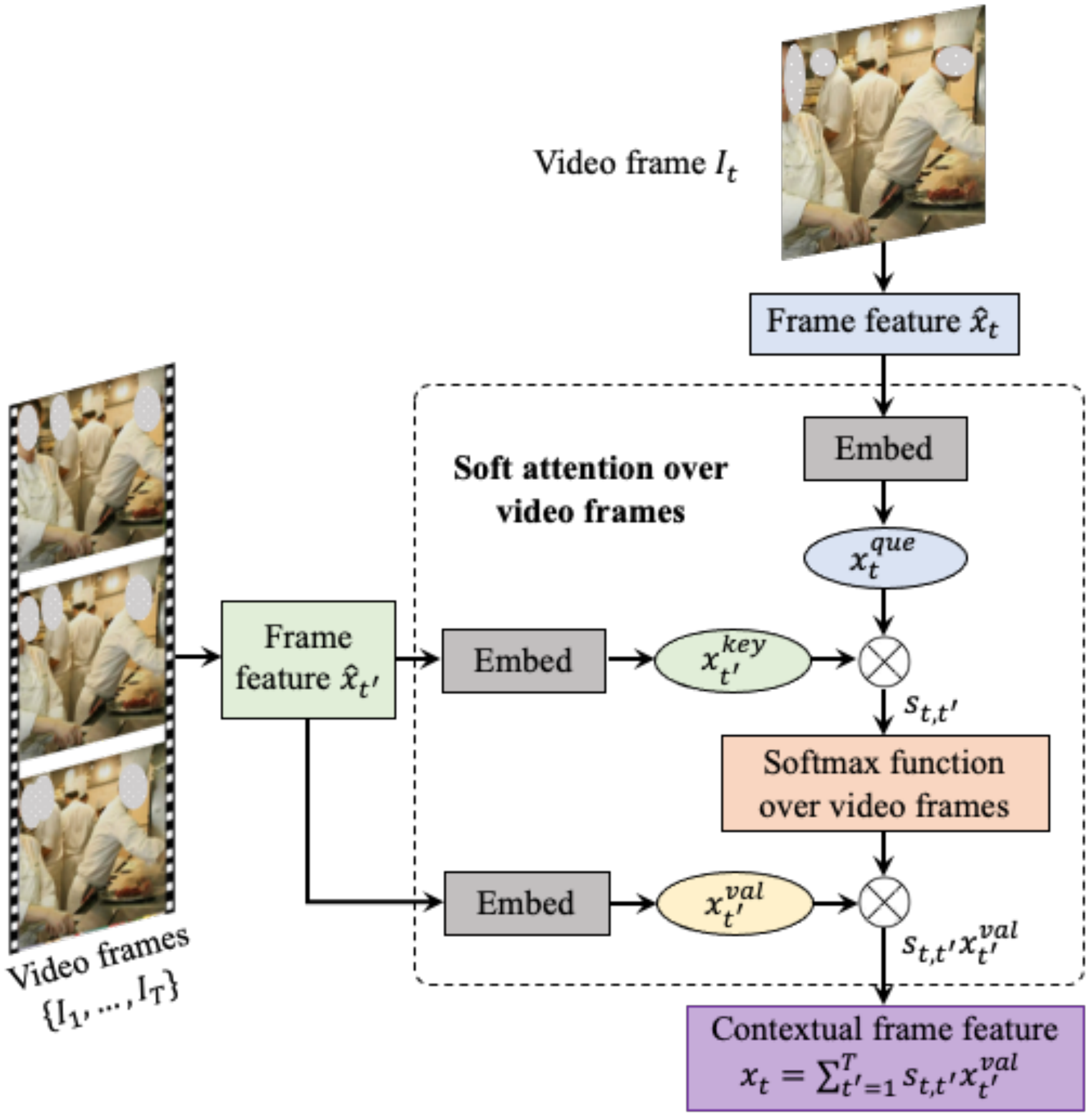}
\end{center}
\vspace{-15pt}
\caption{\small \textbf{Illustration of learning contextual frame feature.} 
Given a frame feature $\hat{x}_t$ obtained by passing this frame through a neural network, we send $\hat{x}_t$ to an embedding layer to generate a ``query'' feature vector $x^{que}_t$. For the feature of each frame in the same video, we use two different embedding layers to get ``key'' $x^{key}_{t'}$ and ``value'' $x^{val}_{t'}$ vectors.
We compute the inner product of the ``query'' and ``key'' to get a similarity score $s_{t,t'}$ of the current frame and each frame in the same video. A softmax layer is then applied to the similarity scores to normalize the similarity of each frame to the current frame. 
The contextual frame feature is obtained by the weighted average over frame ``value'' features.
(Video credit: 
The Best Gallery Craft~\cite{urlfig2})}
\label{apx_fig:temporal_soft_attention}
\vspace{-10pt}
\end{figure}

\textbf{Contextual frame feature learning.}
We describe the contextual frame feature learning in the main paper Section 3.2. Here we illustrate the learning process of the contextual frame feature in \fig{apx_fig:temporal_soft_attention}.

Human-object interactions are temporal events and occur over a period of time. 
To utilize the temporal information from the whole video, we use a soft attention module \cite{vaswani2017attention} to learn a contextual feature representation $x_t$ for each frame.
Given a video frame $I_t$, we send the frame to Faster R-CNN \cite{ren2015faster} and extract the final layer of the FasterR-CNN feature pyramid network to obtain an intermediate feature map. We add an average pooling layer after the intermediate feature map and generate a feature vector as the frame feature descriptor $\hat{x}_t$.
Then we send $\hat{x}_t$ to an embedding layer to generate a ``query'' feature vector $x^{que}_t$. 
We use the same method to extract the features of all frames in the input video and represent them as $\{\hat{x}_1, \cdots, \hat{x}_T\}$.
For the feature of each frame in the video, we use two different embedding layers to get ``key'' $x^{key}_{t'}$ and ``value'' $x^{val}_{t'}$ vectors.
We compute the inner product of the ``query'' and ``key'' to get a similarity score $s_{t,t'}=(x^{que}_t)^T x^{key}_{t'}$ of the current frame and each frame in the same video. A softmax layer is then applied to the similarity scores to normalize the similarity of each frame to the current frame. 
The contextual frame feature is obtained by the weighted average over frame ``value'' features $x_t = \sum_{t'=1}^T {s_{t,t'} x^{val}_{t'}}$.

\begin{figure}[t]
\begin{center}
\includegraphics[width=0.48\textwidth]{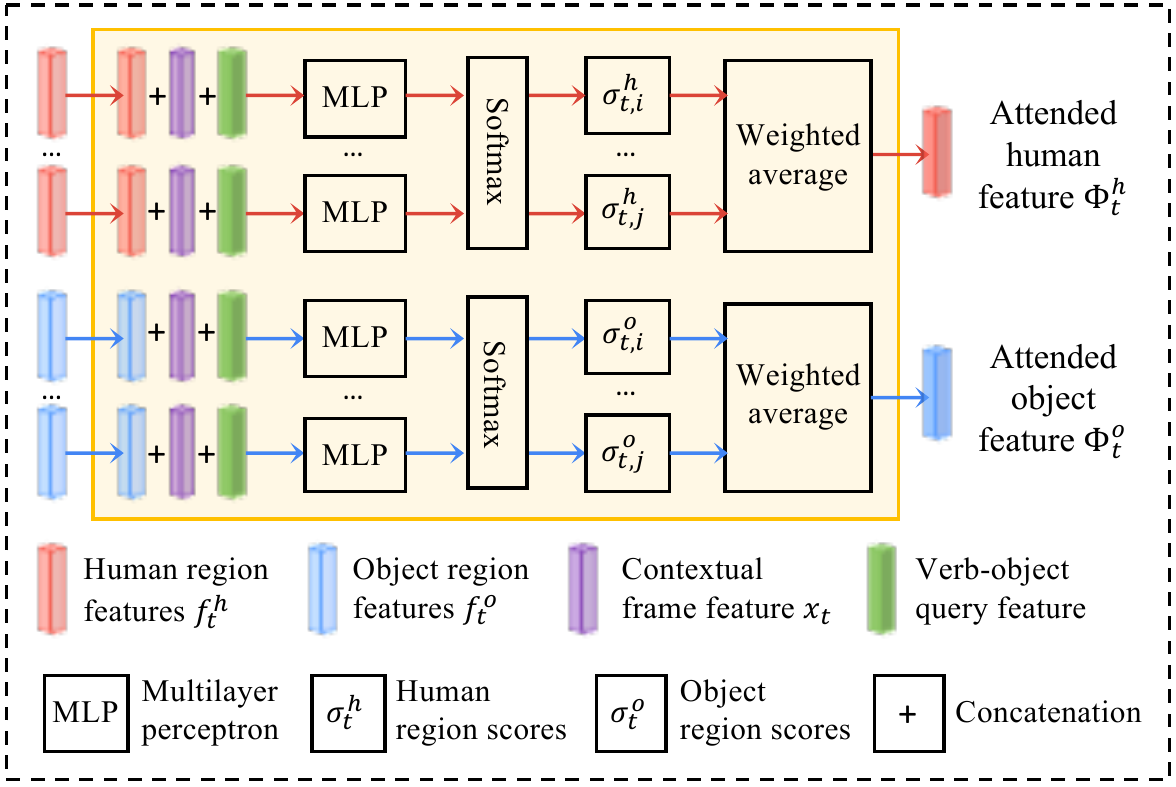}
\end{center}
\vspace{-10pt}
\caption{ \small{ \textbf{Illustration of the region attention module.}} 
The region attention module computes attention scores for the human/object region proposals to measure their relative relevance to the given verb-object query. 
For each human region in frame $I_t$, we first concatenate its feature representation $f_{t,i}^h$ with the contextual frame feature $x_t$ and the verb-object query feature and then pass them through an MLP to obtain a score. We apply the softmax function over the scores of all human regions in this frame and get the final human attention scores $\sigma^h_t$. 
Similarly, each object region has an object attention score $\sigma^o_t$ after applying the softmax function over all object regions.
The attention scores are used to aggregate human/object features as weights in a weighted average given by Equation 2 in the main paper. 
}
\label{apx_fig:attention_model}
\vspace{-15pt}
\end{figure}

\textbf{Region attended human/object feature learning.}
To obtain the attended human and object features, $\Phi_t^h$ and $\Phi_t^o$, used in the main paper Figure 2, we first compute an attention score for each human/object region and then aggregate the human/object features based on their attention scores.
In \fig{apx_fig:attention_model}, we show the details of the region attention module used in the main paper (Figure 2).
The region attention module computes attention scores for the human/object region proposals to measure their relative relevance to the given verb-object query. 
For each human region in frame $I_t$, we first concatenate its feature representation $f_{t,i}^h$ with the contextual frame feature $x_t$ and the verb-object query feature and then pass them through a small network (consisting of two fully-connected layers with LeakyReLU as the activation function in the middle) to obtain a score. We apply the softmax function over the scores of all human regions in this frame and get the final human attention scores $\sigma^h_t$. 
Similarly, each object region has an object attention score $\sigma^o_t$ after applying the softmax function over all object regions.
The attention scores are used to aggregate human/object features using Equation 2 in the main paper.

% \begin{figure}[t]
% \begin{center}
% \includegraphics[width=0.5\textwidth]{fig/files/attention_model.pdf}
% \end{center}
% \caption{ \small{Attention model.}}
% \label{fig:attention_model}
% \end{figure}

% ------------------------------------------------------------------------------------------------------------------------------------------------------------------------------------------------

\subsection{Implementation details}
\label{sect:implementation}
Our model is initialized with a ResNext101 Faster R-CNN model, with the RPN pretrained on the COCO dataset from the Detectron library \cite{Detectron2018}. 
% Our `unseen' test set (51 unseen object classes) has 2 classes present in the COCO object vocabulary, indicating only 2 classes are seen during pre-training.
During training, we select 12 frames from each video and $512$ object region proposals (after non-maximum suppression) as object candidate bounding boxes and $25$ human bounding boxes for each frame. 
For the weakly supervised language-embedding alignment $\losslanguage$ loss (Equation 3 of the main paper), we compute the loss over 15 sampled negatives from $\dictembedding^v$ for the human term and 15 sampled negatives from $\dictembedding^o$ for the object term, during training. 

For the self-supervised temporal contrastive loss $\losstemporal$ in each frame $I_t$, we compute the loss over 15 sampled negatives from the negative feature set $\featuresnegative_t^o$. 
In practice, we find that the objects or humans of interest are not always present across all the frames in a video.
Some video frames will only show part of the object/human or background.
To make the proposed self-supervised temporal contrastive loss more robust to frames that do not contain the mentioned human-object interaction, we only use the temporal contrastive losses on 50\% frames that have the lowest temporal contrastive losses in each video. The selected frames are more likely to contain the target human and objects.

We used the Adam optimizer \cite{Kingma2015Adam} with a learning rate of 1e-4 and a learning rate of 1e-6 for the Faster R-CNN. 
We use a weight coefficient of $\alpha=0.1$ for the temporal contrastive loss $\losstemporal$ in Equation 5 of the main paper.

\vspace{-3pt}
\section{Dataset collection details}
\label{sec:dataset_collection_details}
\vspace{-3pt}
We extract the videos from the Moments in Time dataset \cite{monfort2019moments}.
The Moments in Time dataset has 800k videos with associated metadata, such as title sentences and tags. Moreover, each video has a manually provided action label, such as ``drinking'' and ``pushing''. 
We leverage the action labels to help find labels for the human-object interactions from the metadata associated with the videos. 
We achieve this goal by initially filtering videos to contain the action label in the title sentence or metadata.
However, some videos do not have verbs corresponding to human-object interactions, such as ``storming'' and ``erupting'', so we manually discard videos that do not correspond to human actions.
We then used the Stanford NLP parser \cite{manning2014stanford} to find videos containing noun phrases after the action label in the title or metadata, and use the resulting noun phrase as the object label. 
Finally, we remove videos with non-English metadata and manually filter out bad parsing results. 
After filtering, we obtained approximately 14,000 videos. We manually filtered out bad examples, such as videos having low frame resolution, wrong language labels, or blurry humans and objects.
We finally obtained 6,594 videos in total.

We semi-automatically analyzed the natural language descriptions that accompanied the videos.
% Our dataset construction is different from existing ones in that 
We do not define a fixed list of HOIs a priori but instead use action-object pairs that appear with a certain frequency in the language captions.
By considering more videos with accompanying descriptions, the vocabulary naturally increases.

We collect human and object bounding box annotations using Amazon Mechanical Turk for the test and unseen datasets. We ask each worker to annotate the specific human and object bounding boxes participating in the given human-object interaction label. For each video frame, we collect bounding box annotations from 3 different workers.
We average the annotations from each worker to obtain the object bounding box annotations.
We assume that there can be multiple people interacting with the given object in a video frame. To obtain the accurate number of humans in the input video frame, we want to cluster the human bounding boxes collected from different workers. The close human bounding boxes are more likely to describe the same human.
By counting the number of clusters, we can estimate the number of humans in the input video frame.
To do this, we ran an affinity propagation clustering algorithm \cite{scikit-learn} on all labelled human bounding boxes across multiple workers. We select the clusters which have more than two annotations and average all the annotations within each cluster as the bounding box annotation of that person. 
We further manually examine the annotated bounding boxes and discard low-quality annotations.

\vspace{-3pt}
\section{Dataset statistics}
\label{sec:dataset_statistics}
\vspace{-3pt}
Our focus is on video-based, human-centric HOI detection without exhaustively annotating the spatial location of objects in a video at training which is time consuming given the large number of frames in a video.
Our dataset consists of 244 different object classes and 99 different action classes.
There are 756 verb-object classes in total with diverse human-object interactions.

All the videos are extracted from the Moments in Time dataset \cite{monfort2019moments}, which contains short trimmed videos. 
% The action-object vocabulary is open in the sense that we semi-automatically analyzed the natural language descriptions that accompanied the videos.
We semi-automatically analyzed the natural language descriptions that accompanied the videos.
% Our dataset construction is different from existing ones in that we do not define a fixed list of HOIs a priori but instead use action-object pairs that appear with a certain frequency in language captions.
By considering more videos with accompanying descriptions, the vocabulary can naturally increase.

% We provide the qualitative video frames of our dataset in \fig{fig:dataset}.
% of this supplement.

\begin{figure*}
\begin{center}
\includegraphics[width=1\linewidth]{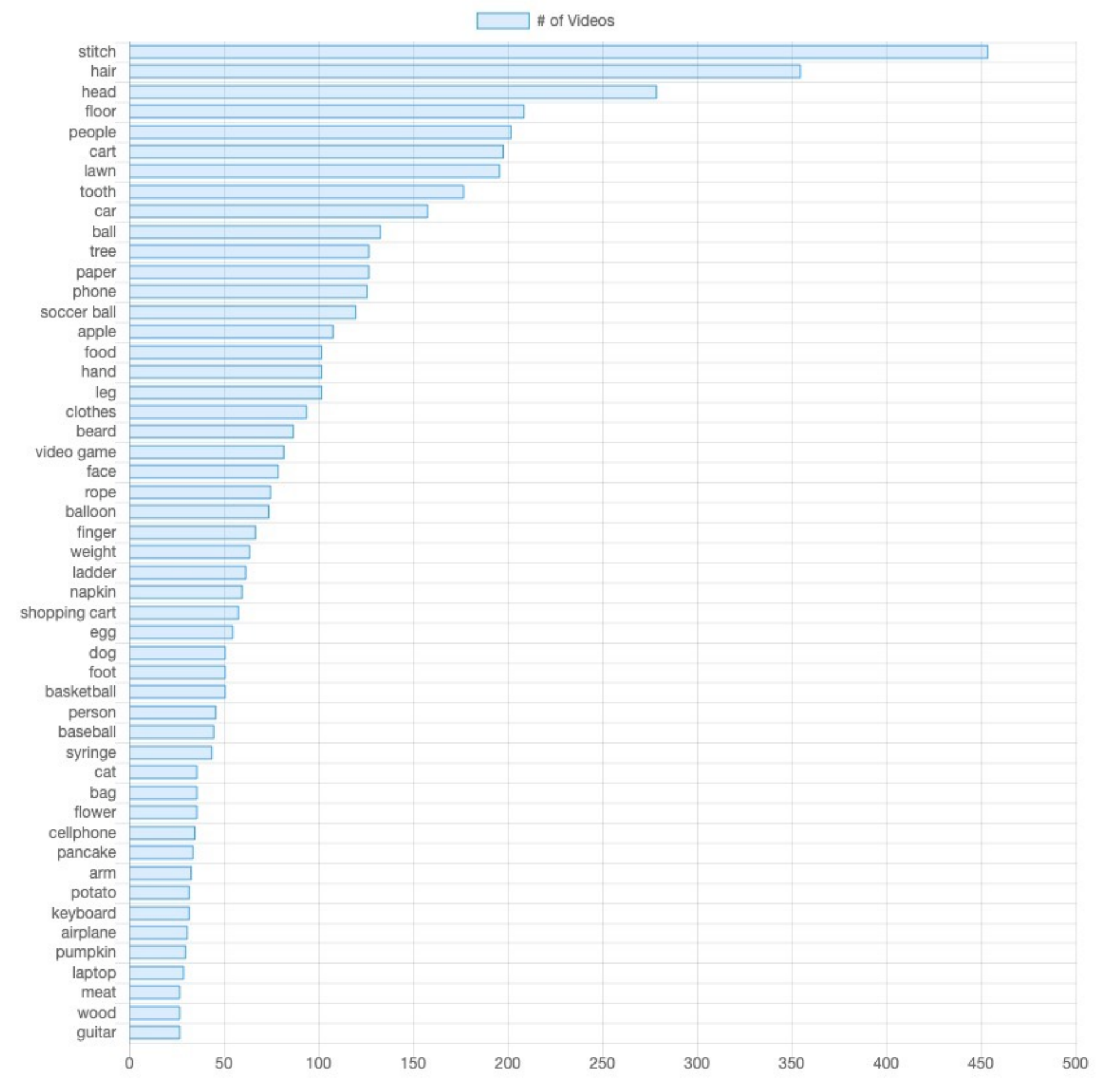}
\end{center}
\vspace{-15pt}
\caption{\textbf{Distribution of objects in our dataset.} Our dataset consists of 244 different object classes, where for brevity we only show the top 50 in the diagram above.}
\label{apx_fig:obj}
\end{figure*}

\begin{figure*}
\begin{center}
\includegraphics[width=1\linewidth]{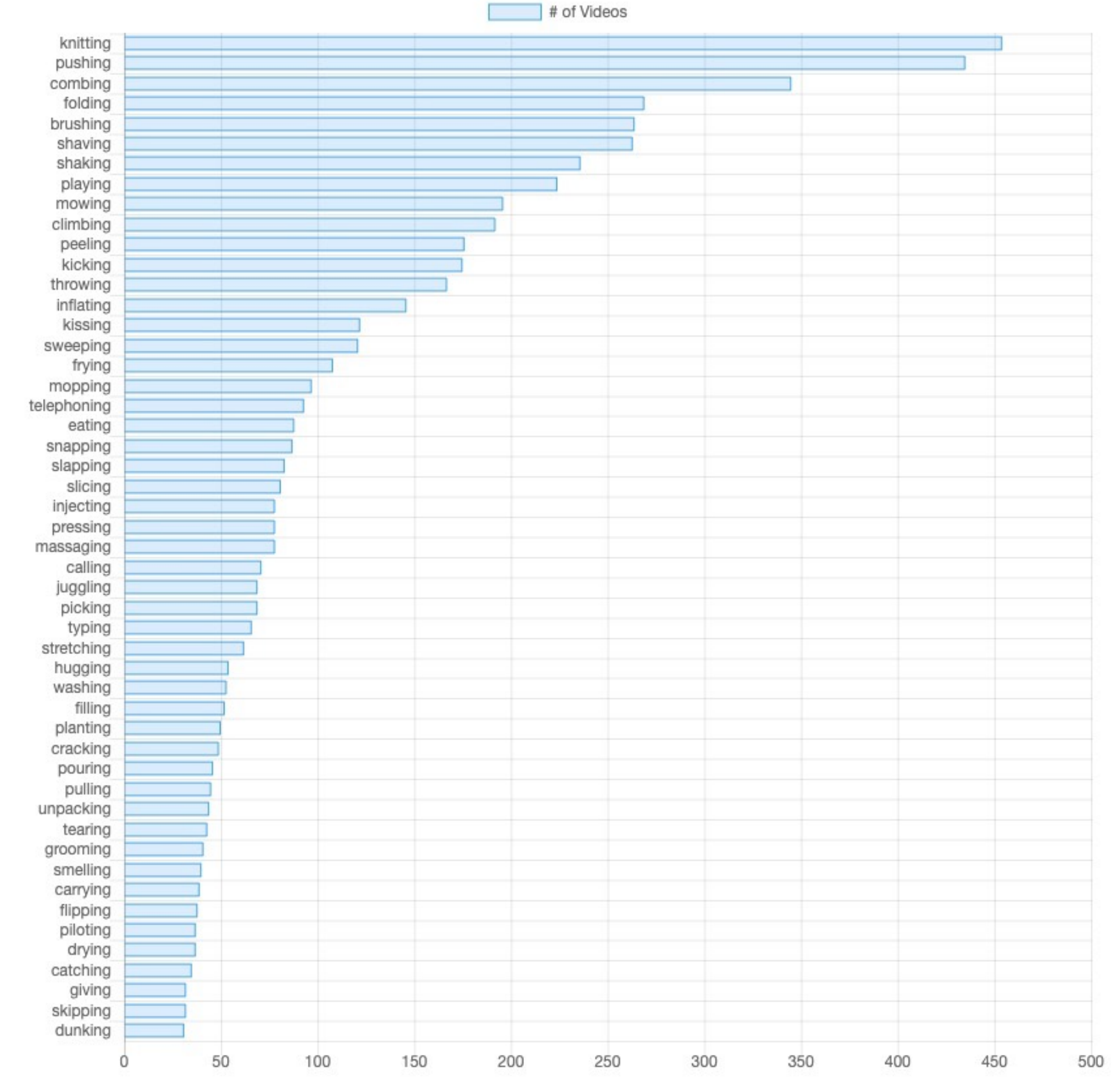}
\end{center}
\vspace{-15pt}
\caption{\textbf{Distribution of actions in our dataset.} Our dataset consists of 99 different action classes, where for brevity we only show the top 50 in the diagram above.}
\label{apx_fig:action}
\end{figure*}

\begin{figure*}
\begin{center}
\includegraphics[width=1\linewidth]{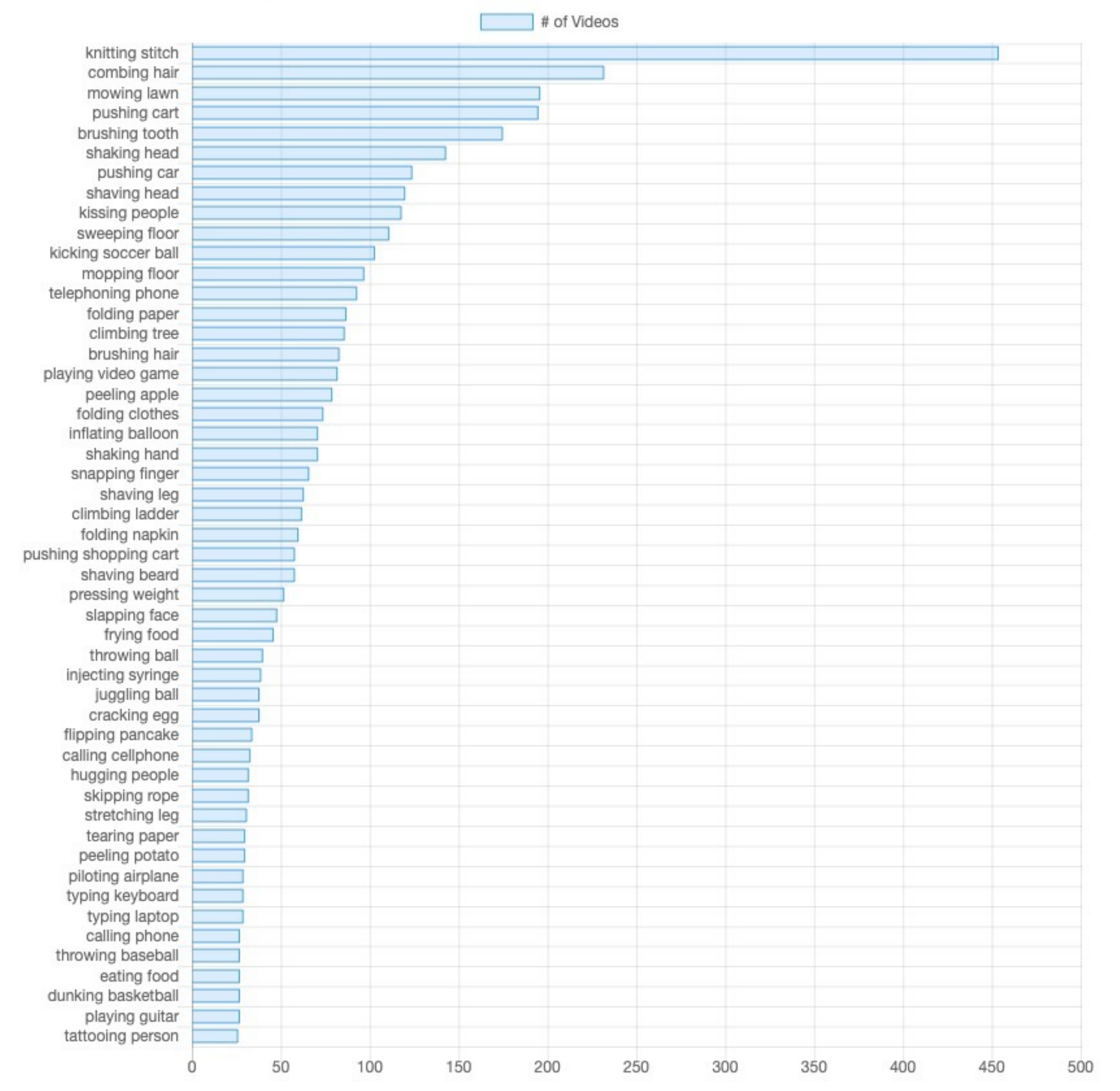}
\end{center}
\vspace{-15pt}
\caption{\textbf{Distribution of verb-object classes in our dataset.} Our dataset consists of 756 different verb-object classes, where for brevity we only show the top 50 in the diagram above.}
\label{apx_fig:triplet}
\end{figure*}

We present the dataset statistics in \fig{apx_fig:obj}, \fig{apx_fig:action}, and \fig{apx_fig:triplet}.
\fig{apx_fig:obj} shows the distribution of objects. We show the top 50 most frequent object classes. 
\fig{apx_fig:action} shows the distribution of the top 50 most frequent action classes.
\fig{apx_fig:triplet} shows the distribution of the top 50 most frequent verb-object classes.

% \begin{figure*}
% \begin{center}
% \includegraphics[width=1\linewidth]{fig/files/dataset_examples2.pdf}
% \end{center}
% \vspace{-15pt}
% \caption{\textbf{Example videos from \lsdata with verb-object queries below.} Our video data set consists of diverse human object interactions.}
% \label{fig:dataset}
% \end{figure*}

\clearpage

{\small
\bibliographystyle{ieee_fullname}
\bibliography{egbib, reference}
}

\end{document}